\documentclass[11pt, a4paper, logo, copyright]{main}

\usepackage{microtype}
\usepackage{graphicx}
\usepackage{subcaption}
\usepackage{booktabs}
\usepackage{hyperref}
\usepackage{multirow}
\usepackage{colortbl}
\usepackage{xcolor}
\usepackage[most]{tcolorbox}
\usepackage{fvextra}

\usepackage[utf8]{inputenc}
\usepackage[T1]{fontenc}

\usepackage{amsmath, amssymb, amsfonts}

\usepackage{graphicx}
\usepackage{wrapfig}
\usepackage{subcaption}
\usepackage[figuresleft]{rotating}

\usepackage{booktabs}
\usepackage{array}
\usepackage{multirow}
\usepackage{makecell}
\usepackage{longtable}
\usepackage{tabularray}
\usepackage{colortbl}
\usepackage[table]{xcolor}

\usepackage{xcolor}
\usepackage{color}

\usepackage{hyperref}
\usepackage{url}

\usepackage{microtype}
\usepackage{nicefrac}

\usepackage{enumitem}

\usepackage{algorithm}
\usepackage{algorithmic}

\usepackage{listings}
\usepackage{minted} 

\usepackage{tcolorbox}
\tcbuselibrary{breakable, skins}

\usepackage{fontawesome5}
\usepackage{adjustbox}
\usepackage{multicol}
\usepackage{lipsum}

\usepackage[tableposition=top]{caption}

\definecolor{firstcolor}{HTML}{F49767}  
\definecolor{secondcolor}{HTML}{F9BA86} 
\definecolor{thirdcolor}{HTML}{F5D5AC}  
\definecolor{goldcolor}{HTML}{F49767}
\definecolor{silvercolor}{HTML}{F5D5AC}
\definecolor{checkcolor}{RGB}{0,170,0}
\definecolor{xcolor}{RGB}{255,0,0}
\definecolor{synthcol}{RGB}{242, 247, 255}

\definecolor{lightpink}{RGB}{255, 230, 240}
\definecolor{lightblue}{RGB}{230, 240, 255}
\definecolor{lightgray}{RGB}{248, 248, 248}
\definecolor{bordergray}{RGB}{200, 200, 200}

\definecolor{bestcolor}{RGB}{219, 208, 237}
\definecolor{secondcolor}{RGB}{241, 237, 248}
\definecolor{thirdcolor}{RGB}{211, 222, 190}

\definecolor{line-blue}{RGB}{243, 248, 252}
\definecolor{ForestGreen}{RGB}{34, 139, 34}

\definecolor{lightblue2}{rgb}{0.22,0.45,0.70}
\definecolor{lightgreen}{rgb}{0.1, 0.6, 0.1}

\usepackage{pifont}

\newcolumntype{C}[1]{>{\centering\arraybackslash}p{#1}}
\newcolumntype{L}[1]{>{\raggedright\arraybackslash}p{#1}}

\newtcolorbox{promptbox}[2][Prompt]{
  colback=black!5!white,
  arc=5pt,
  boxrule=0.5pt,
  fonttitle=\bfseries,
  title=#1,
  before upper={\small},
  fontupper=\fontfamily{ptm}\selectfont,
  colframe=#2,
}

\usepackage{amsmath}
\usepackage{amssymb}
\usepackage{mathtools}
\usepackage{amsthm}

\theoremstyle{plain}

\theoremstyle{definition}

\theoremstyle{remark}

\usepackage[authoryear, sort&compress, round]{natbib}
\let\cite\citep
\usepackage{float}

\title{SpatialEvo: Self-Evolving Spatial Intelligence via Deterministic Geometric Environments}

\author[*]{
\textbf{Dingming Li}$^{1,*}$, \textbf{Yingxiu Zhao}$^{2,*}$, \textbf{Xinrui Cheng}$^{1}$, \textbf{Kangheng Lin}$^{2}$, \\
\textbf{Hongbo Peng}$^{2}$, \textbf{Hongxing Li}$^{1}$, \textbf{Zixuan Wang}$^{1}$, \textbf{Yuhong Dai}$^{2}$, \\
\textbf{Haodong Li}$^{2}$, \textbf{Jia Wang}$^{2}$, \textbf{Yukang Shi}$^{2}$, \textbf{Liang Zhao}$^{2}$, \textbf{Jianjian Sun}$^{2}$, \textbf{Zheng Ge}$^{2}$, \\
\textbf{Xiangyu Zhang}$^{2}$, \textbf{Weiming Lu}$^{1}$, \textbf{Jun Xiao}$^{1}$, \textbf{Yueting Zhuang}$^{1}$, \textbf{Yongliang Shen}$^{1,\dag}$

$^{1}$ Zhejiang University \quad $^{2}$ StepFun \\
\hspace{1.0em}
\raisebox{-0.15em}{\includegraphics[height=0.9em]{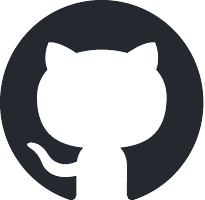}}\hspace{0.35em}\href{https://github.com/ZJU-REAL/SpatialEvo}{GitHub}
\hspace{1.0em}
\raisebox{-0.15em}{\includegraphics[height=0.9em]{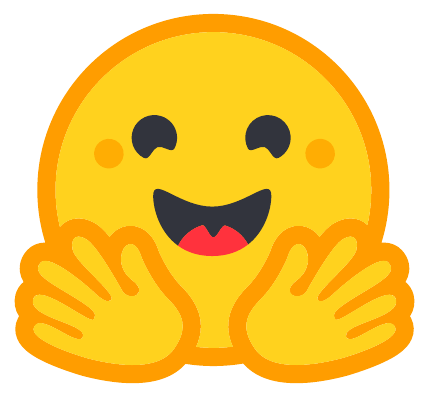}}\hspace{0.35em}\href{https://huggingface.co/lidingm/SpatialEvo-3B}{Hugging Face}
}

\setauthorfootnote{$*$ Equal Contribution \quad $\dag$ Corresponding Author}

\reportnumber{} %

\renewcommand{\today}{}

\begin{abstract}

\noindent
Spatial reasoning over three-dimensional scenes is a core capability for embodied intelligence, yet continuous model improvement remains bottlenecked by the cost of geometric annotation. The self-evolving paradigm offers a promising path, but its reliance on model consensus to construct pseudo-labels causes training to reinforce rather than correct the model's own geometric errors. We identify a property unique to 3D spatial reasoning that circumvents this limitation: ground truth is a deterministic consequence of the underlying geometry, computable exactly from point clouds and camera poses without any model involvement. Building on this insight, we present SpatialEvo, a self-evolving framework for 3D spatial reasoning, centered on the Deterministic Geometric Environment (DGE). The DGE formalizes 16 spatial reasoning task categories under explicit geometric validation rules and converts unannotated 3D scenes into zero-noise interactive oracles, replacing model consensus with objective physical feedback. A single shared-parameter policy co-evolves across questioner and solver roles under DGE constraints: the questioner generates physically valid spatial questions grounded in scene observations, while the solver derives precise answers against DGE-verified ground truth. A task-adaptive scheduler endogenously concentrates training on the model's weakest categories, producing a dynamic curriculum without manual design. Experiments across nine benchmarks demonstrate that SpatialEvo achieves the highest average score at both 3B and 7B scales, with consistent gains on spatial reasoning benchmarks and no degradation on general visual understanding.

\end{abstract}

\begin{document}

\maketitle

\definecolor{colorfirst}{RGB}{252,141,89}
\definecolor{colorsecond}{RGB}{253,187,132}
\definecolor{colorthird}{RGB}{253,212,158}
\definecolor{colorfourth}{RGB}{254,232,200}
\definecolor{colorfifth}{RGB}{255,247,236}
\definecolor{myred}{RGB}{242,128,128}
\definecolor{mygreen}{RGB}{112,180,143}
\definecolor{myblue}{RGB}{210,225,255}
\definecolor{citypink}{RGB}{227,108,194}
\definecolor{cityblue}{RGB}{128,159,225}

\newcommand{\ph}[1]{\textcolor{black}{#1}}
\newcommand{\rankfirst}[0]{\cellcolor{colorfirst}}
\newcommand{\ranksecond}[0]{\cellcolor{colorsecond}}
\newcommand{\rankthird}[0]{\cellcolor{colorthird}}
\newcommand{\rankfourth}[0]{\cellcolor{colorfourth}}
\newcommand{\rankfifth}[0]{\cellcolor{colorfifth}}
\DeclareRobustCommand{\legendsquare}[1]{%
  \textcolor{#1}{\rule{2ex}{2ex}}%
}

\section{Introduction}
\label{intro}
Spatial reasoning requires models to form persistent perception and understanding of complex three-dimensional scenes, serving as a core capability for embodied intelligence, robot navigation, and scene question answering~\cite{chen2024spatialvlm, cheng2024spatialrgpt, li2025embodied, ma2024llms}. 
Yet how to continuously drive model iteration at scale and overcome the bottleneck of data acquisition remains an open challenge, one that existing approaches address primarily through large-scale annotated datasets pairing multi-view images with geometric question-answer pairs~\cite{chen2024spatialvlm, song2025robospatial}.

These efforts have yielded meaningful progress, yet they share a common structural weakness: the training distribution is fixed at dataset creation time. A static corpus cannot respond to where a model is weak today, cannot generate harder examples as the model grows stronger~\cite{wang2025spatial457, tian2025nuscenes}, and cannot scale without proportional investment in human annotation~\cite{zheng2026pearl, lyu2024mmscan}. This raises a fundamental question: \textit{how can a model continuously improve its spatial reasoning without depending on an ever-growing supply of labeled data?}

The self-evolving paradigm offers a principled answer. By actively distilling training signals from raw visual inputs through iterative self-play, self-evolving models can progressively internalize reasoning paradigms~\cite{he2025visplay, deng2024enhancing, li2025mixture} and continuously update their training distribution alongside their own capability, providing a natural dynamic curriculum without manual curation~\cite{liu2024diving, zhao2025absolute}. Yet existing self-evolving methods share a critical limitation: because ground truth cannot be read off the environment directly, the training signal must be constructed by aggregating the model's own predictions through majority voting or self-consistency\cite{tao2024survey, wu2024continual, shi2025continual, wang2025language, he2025visplay}. This introduces a systematic bias: the pseudo-labels inherit the model's own prediction errors, and gradient updates anchored to such labels risk reinforcing rather than correcting the model's existing errors rather than correcting them.

We observe that 3D spatial reasoning possesses a distinctive property that circumvents this limitation entirely. Unlike natural language or general vision tasks, the ground truth for a spatial question is a deterministic consequence of the underlying geometry~\cite{luo2025geogrambench, ma20253dsrbench, xu2025spatialbench, cheng2024spatialrgpt}. Given a dense point cloud, calibrated camera poses, and a well-formed geometric question, the correct answer can be computed exactly and programmatically, with no appeal to any model's judgment. 

A question about absolute distance reduces to a nearest-point computation on object bounding boxes; a question about relative camera orientation reduces to an arithmetic operation on rotation matrices~\cite{mayer2025ivispar, zhou2025vlm4d}. The model consensus that other domains rely on as a noisy proxy is therefore unnecessary~\cite{song2025robospatial, cai2025spatialbot}: the physical world itself serves as an exact and impartial judge. Every unannotated 3D indoor scene is, in principle, an inexhaustible source of noise-free supervision~\cite{liu2025ssr, cheng2024spatialrgpt}, waiting to be converted into a training signal.

\begin{figure}[t]
\centering
\includegraphics[width=0.98\textwidth,
    trim={55pt 0pt 10pt 0pt},
    clip]{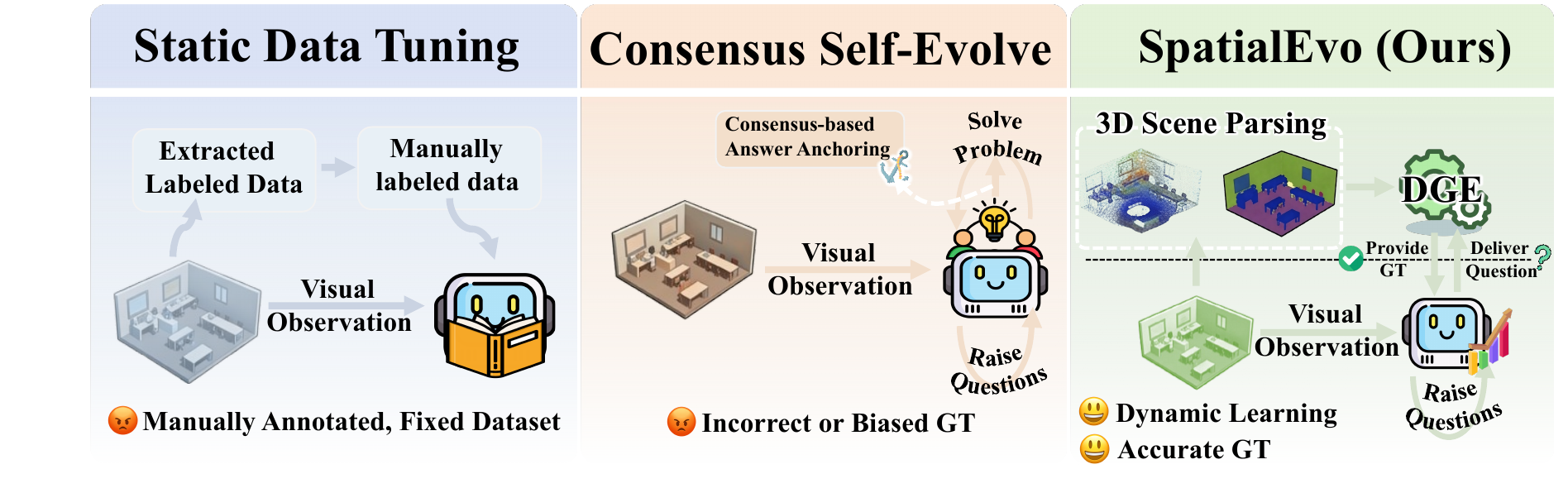}
\caption{Comparison of three training paradigms. Static Data Tuning relies on manually annotated fixed datasets. Consensus Self-Evolve eliminates manual annotation but introduces biased ground truth via model voting. SpatialEvo (Ours) leverages the DGE to compute exact ground truth from 3D scene assets, enabling zero-noise dynamic curriculum learning.}
\label{figure:intro}
\vspace{-1.2em}
\end{figure}
To this end, we propose \textbf{SpatialEvo}, the first framework to introduce the self-evolving paradigm into 3D spatial reasoning. Its core contribution is the \textbf{Deterministic Geometric Environment (DGE)}, which designs an atomic geometric verification rule set covering 16 spatial reasoning task categories, programmatically computes exact ground truth from 3D point clouds and camera pose sequences, and transforms 3D scene assets into zero-noise online reward judges~\cite{cheng2024spatialrgpt}. Building on this, the GRPO~\cite{shao2024deepseekmath} framework drives a single VLM to co-evolve between questioner and solver roles~\cite{he2025visplay, li2026mm}: the questioner perceives the overall scene layout and generates physically valid spatial questions, while the solver internalizes spatial reasoning paradigms through explicit geometric derivation under DGE constraints. This division naturally aligns with the two cognitive levels of spatial reasoning, namely holistic environment perception and local geometric solving. A task scheduler dynamically regulates the sampling distribution based on historical accuracy, yielding adaptive curriculum learning~\cite{shi2025efficient}. 

Extensive experiments across nine benchmarks show that SpatialEvo achieves the highest average score at both 3B and 7B scales, with consistent gains on spatial reasoning benchmarks and no degradation on general visual understanding. Ablation studies confirm that replacing DGE ground truth with majority-vote pseudo-labels produces the single largest performance drop, directly validating the role of deterministic physical feedback.

In summary, our contributions are:

\begin{itemize}
    \item We propose \textbf{SpatialEvo}, a self-evolving framework for 3D spatial reasoning centered on a deterministic geometric environment, integrating VLM self-play with programmatic physical verification to replace model consensus with objective environmental feedback for noise-free continuous evolution.
    \item We design an automated geometric verification pipeline that decouples 3D spatial reasoning into executable atomic verification rules spanning 16 task categories, transforming unannotated 3D scene datasets into an online-interactive, zero-noise ground truth judge engine.
    \item We introduce a spatially grounded policy co-evolution mechanism where a single model simultaneously occupies questioner and solver roles under DGE constraints, with adaptive task scheduling driving curriculum self-emergence, validated through extensive experiments.
\end{itemize}

\section{Related Works}
\paragraph{Spatial Reasoning in Vision-Language Models}
Early work sought to enhance the spatial perception of VLMs through multi-view image fusion or depth information injection~\cite{chen2024spatialvlm, cheng2024spatialrgpt, ma2024spatialpin}. More recent approaches have proceeded along two parallel tracks: data construction and reasoning-oriented training. Methods such as SpatialVLM~\cite{chen2024spatialvlm}, SpatialBot~\cite{cai2025spatialbot}, and SpatialLadder~\cite{li2025spatialladder} achieve notable gains by fine-tuning on large-scale spatial annotation datasets~\cite{cai2025scaling, ouyang2025spacer}, while other efforts explore programmatic synthesis or bird's-eye-view augmentation to reduce annotation costs~\cite{yang2025visual}. However, the datasets produced by these approaches are static by nature, fixed at generation time and thus insensitive to the model's dynamically shifting weaknesses during training, which fundamentally limits training efficiency.

\paragraph{Self-Evolution of Vision-Language Models}
Self-evolution has emerged as a prominent research direction in the LLM community~\cite{deng2025self, li2025self, zhai2024fine, tao2024survey} and has progressively extended to VLMs~\cite{liu2024diving, deng2024enhancing}. Notably, the spatial reasoning domain is particularly well-suited for self-evolution, as its visual inputs inherently carry physical information enabling deterministic ground truth computation, naturally circumventing reliance on model consensus. VisPlay~\cite{he2025visplay}, EvolMM~\cite{thawakar2025evolmm}, and V-Zero~\cite{wang2026v} advance multimodal self-play but remain constrained by static image corpora; Vision-Zero~\cite{wang2025vision} and MM-Zero~\cite{li2026mm} achieve continual improvement via verifiable rewards or zero-data rendering. Critically, all rely on model consensus as a reward proxy, introducing systematic bias where precise physical grounding is required. SpatialEvo is the first to bring self-evolution to 3D spatial reasoning, replacing model-derived judgment with deterministic geometric computation.

\section{Methodology}

\begin{figure}[t]
\centering
\includegraphics[width=0.98\textwidth,
    trim={0pt 0pt 0pt 0pt},
    clip]{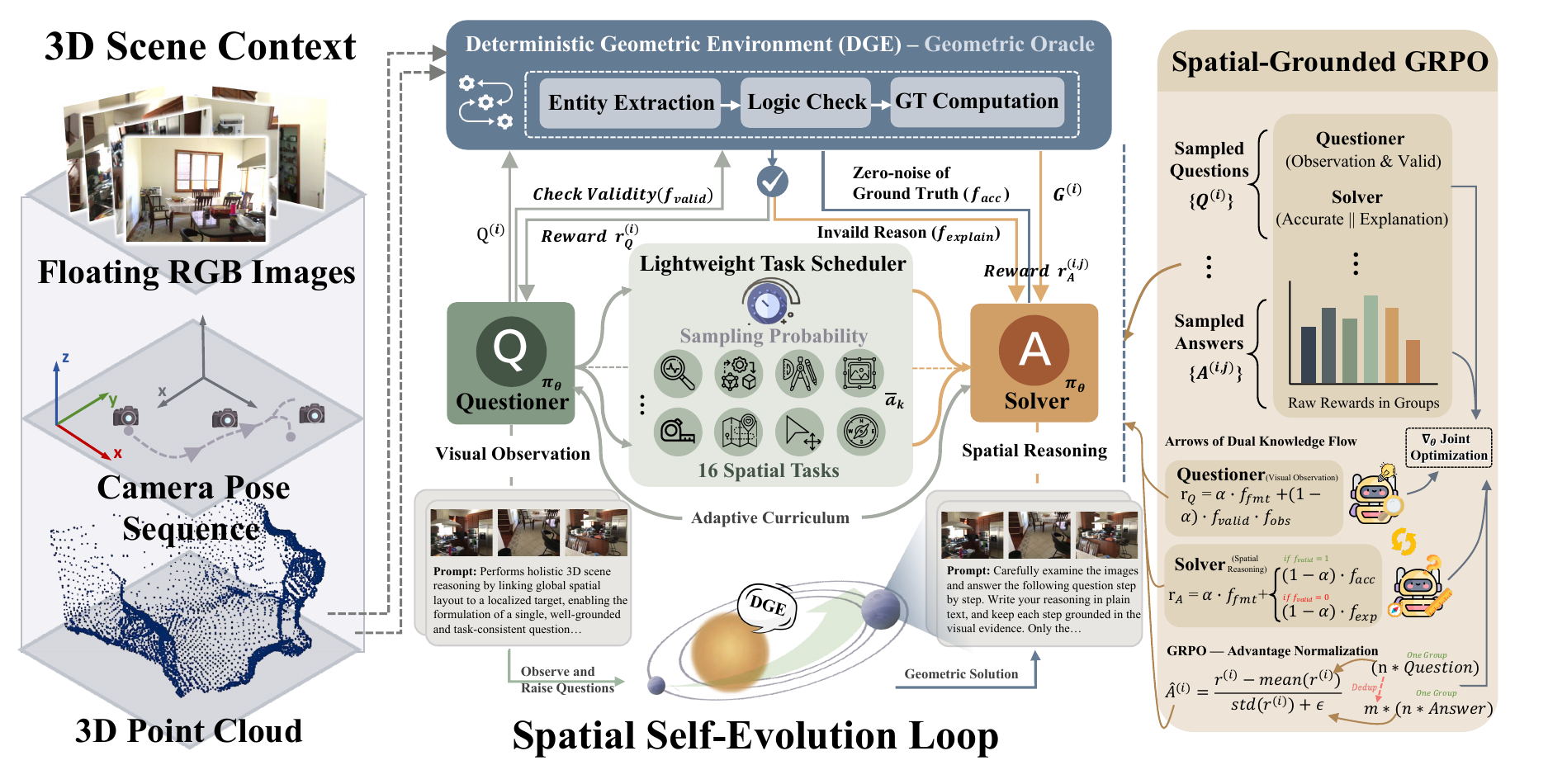}
\caption{Overview of \textbf{SpatialEvo}. The DGE computes zero-noise ground truth directly from 3D point clouds and camera pose sequences, enabling the Spatial Self-Evolution Loop to jointly optimize a single VLM as both Questioner and Solver via Spatial-Grounded GRPO.}
\label{fig:main}
\vspace{-1.2em}
\end{figure}

\subsection{Preliminaries}
\label{Preliminaries}
\paragraph{Problem Formulation.}
We formulate spatial reasoning as a visual question-answering process grounded in multi-view RGB imagery. Given a 3D indoor scene $\mathcal{S}$ whose underlying assets comprise dense point clouds, semantic annotations, and camera pose sequences, the observable input to the policy model is restricted to the corresponding set of RGB image frames $\mathcal{I} = \{I_1, I_2, \ldots, I_T\}$. Our objective is to learn a single vision-language policy model $\pi_\theta$ endowed with two complementary capabilities: as a \textit{questioner}, generating spatially meaningful reasoning questions $Q$ from visual observations; and as a \textit{solver}, predicting precise answers $A$ given an image context and a question description.

The task space $\mathcal{T}$ encompasses 16 core spatial reasoning tasks, organized into three categories by observational granularity. \textit{Multi-image scene-level tasks} (6 tasks)---object counting, object size, absolute distance, relative distance, relative direction, and room size estimation---require the model to integrate global 3D layout information across multiple frames~\cite{yang2025thinking}. \textit{Single-image tasks} (3 tasks)---single-view relative direction, camera-to-object distance (Distance Cam-Obj), and inter-object depth ordering (Depth Order Obj-Obj)---assess the model's understanding of single-frame perspective geometry. \textit{Dual-image tasks} (7 tasks)---inter-camera relative position (Position Cam-Cam), camera-to-object position (Position Cam-Obj), camera-to-region position (Position Cam-Reg), camera motion estimation~\cite{yang2025mmsi}, visibility comparison, inter-camera elevation (Elevation Cam-Cam), and attribute measurement---focus on geometrically consistent inference across viewpoints. The complete definitions and verification rules for all 16 task categories are provided in Appendix~\ref{app:B.1}.

Collectively, these tasks span three semantic dimensions: metric measurement, topological spatial relations, and camera pose reasoning. For instance, computing the relative orientation between two cameras requires resolving the rotational component between their extrinsic matrices, while inferring inter-object depth ordering requires aligning perspective projection with the scene point cloud. A defining characteristic shared by all tasks is that their answers can be computed programmatically and exactly from the scene's underlying geometric assets, providing a natural foundation for automatic ground-truth generation and, in turn, for noise-free reinforcement learning signals.

\paragraph{Framework Overview.}
As illustrated in Figure~\ref{fig:main}, the SpatialEvo framework consists of two core components: the Deterministic Geometric Environment (DGE) and Spatial-Grounded Policy Co-Evolution. The DGE takes 3D point clouds and camera pose sequences as underlying scene assets, applies task-specific geometric verification rules to validate candidate questions, and programmatically computes exact ground-truth for verified questions, serving as a noise-free physical feedback judge. Policy co-evolution is built on the GRPO algorithm, where a single policy model $\pi_\theta$ switches between Questioner and Solver via role-conditioned prompting: the Questioner perceives the global 3D scene layout to generate spatially valid and reasoning-worthy questions, while the Solver focuses on precise geometric inference under the hard constraints of DGE-verified ground truth, with rewards anchored directly to physical computation results. A lightweight task scheduler dynamically adjusts the task sampling distribution based on the Solver's historical performance, driving the emergence of an adaptive curriculum without any human intervention.

\subsection{Deterministic Geometric Environment}
\label{Deterministic Geometric Environment}
The Deterministic Geometric Environment (DGE) constitutes the physical feedback core of the SpatialEvo framework, serving as the \textit{Geometric Oracle} within the self-evolution loop. Its fundamental function is to map natural language questions generated by the policy model onto the underlying 3D scene assets for objective verification, replacing unreliable model judgment with programmatic geometric computation to supply reinforcement learning with zero-noise supervisory signals. The DGE operates through two tightly coupled submodules: a task-specific geometric validation rule set and an automated verification pipeline. Owing to a rich library of spatial numerical computation tools, the DGE is designed for extensibility: supporting a new task requires only the specification of the corresponding rule set and validation logic.

\subsubsection{Task-Specific Geometric Validation Rule Sets}
For each of the 16 spatial reasoning task categories, the DGE pre-defines a geometric verification rule set that decouples complex spatial intuition into executable atomic criteria, rendering abstract geometric reasoning problems mathematically decidable. Each rule set constrains question validity along three dimensions: \textit{premise consistency} requires that all scene entities referenced in a question (e.g., frame indices, object categories, or spatial regions) exist in the underlying assets and can be uniquely localized; \textit{Inferential solvability} requires that the geometric premises of a question are unambiguously computable: for instance, metric tasks must ensure that the reference object's point cloud density meets a minimum precision threshold, while orientation tasks must ensure sufficient viewpoint disparity between the two frames to constitute a meaningful spatial relationship; \textit{geometric degeneracy filtering} discards physically unstable, highly ambiguous, or low-training-value edge cases. Note that numerical tolerances for continuous-valued tasks are not part of the validity rules but are defined separately at the reward stage; see Appendix~\ref{app:B.1} for this distinction.

Taking camera orientation tasks as an example, the rule set requires the Questioner to specify valid frame index pairs; the DGE then computes the relative rotation matrix and translation vector between the two poses to determine whether they constitute an answerable spatial relationship, mapping the result to a discrete directional label. For object size or room metric tasks, the DGE directly performs bounding box fitting or plane normal estimation on the scene point cloud to obtain precise metric values. For depth ordering tasks, the DGE projects the scene point cloud onto the corresponding camera image plane and derives the ground-truth ordering by comparing the median depth of target objects. This rule-based structured decomposition strategy ensures that the verification logic across all task categories carries unambiguous mathematical semantics. Complete rule definitions and verification logic for all task categories are provided in Appendix~\ref{app:B.1}.

\subsubsection{Automated Verification Pipeline}
Driven by the validation rule sets, the DGE instantiates a fully automated, end-to-end pipeline spanning question parsing to ground-truth synthesis, converting existing 3D scene datasets such as ScanNet~\cite{dai2017scannet} and ScanNet++~\cite{yeshwanth2023scannet++} into interactive, deterministic ground-truth judging engines. The pipeline executes in three stages.

\paragraph{Stage 1: Entity Parsing.}  Since questions generated by the Questioner are in free-form natural language, the DGE invokes a lightweight LLM to perform structured entity extraction over the question text, identifying core semantic elements such as frame indices, object categories, and spatial relationship descriptions, and normalizing them into structured representations directly consumable by the backend geometric computation module. The prompt design is detailed in Appendix~\ref{app:B.2}.

\paragraph{Stage 2: Legality Verification.} The DGE validates the extracted entities against the rule set corresponding to the current task category, checking premise consistency and reasoning chain validity in sequence. If a question references a non-existent scene entity or its geometric premises violate physical constraints, such as an invalid frame index or a reference object whose point cloud is too sparse, the DGE triggers a truncation mechanism and returns a negative reward to the questioner, suppressing the propagation of hallucinated questions at the data generation stage and thereby preserving the overall signal-to-noise ratio of the training signal.

\paragraph{Stage 3: Ground-Truth Synthesis.} For questions that pass legality verification, the DGE invokes its geometric toolkit to perform precise numerical computation in the global coordinate frame. Core operators include rigid-body coordinate transformation, point cloud bounding-box fitting and topological analysis, depth-map perspective projection, and planar normal estimation, collectively covering the full computational requirements of all 16 task categories across the metric, topological, and orientation semantic dimensions. The pipeline outputs not only ground-truth answers with guaranteed precision but also intermediate geometric states recorded during inference, such as keypoint coordinates and camera trajectory segments, providing interpretable intermediate representations for subsequent qualitative analysis and error attribution.

This paradigm of replacing black-box model prediction with programmatic physical computation eliminates reward noise from the self-evolution process, ensuring that every gradient update of the policy model is anchored to objective physical laws.

\subsection{Spatial-Grounded Policy Co-Evolution}
\label{Spatial-Grounded Policy Co-Evolution}
Within the SpatialEvo framework, the co-evolution of the questioner and the solver serves as the primary driver of spatial intelligence advancement. This section details the spatial-grounded policy co-evolution mechanism, covering the self-play architecture, task scheduling design, reward function formulation, and the GRPO training procedure.

\subsubsection{Spatial Self-Play Mechanism}
SpatialEvo employs a single policy model $\pi_\theta$ that alternates between the questioner and solver roles via role-conditioned prompting. Parameter sharing confers a dual advantage: gradients acquired by the solver through geometric derivation directly improve the questioner's visual-spatial perception, while the geometric intuitions developed by the questioner during boundary exploration in turn deepen the solver's reasoning capacity, establishing a virtuous cycle of mutual knowledge reinforcement between the two roles~\cite{he2025visplay, li2026mm}.

In this mechanism, the task scheduler first infers the feasible task set $\mathcal{T}_s^{\text{feasible}}$ from the asset characteristics of the current scene, then samples a task type according to the solver's historical effective accuracy and assigns it to the questioner. The questioner perceives the holistic 3D layout of the scene from multi-view RGB images and generates a physically valid spatial reasoning question for the designated task type. The DGE immediately performs legality verification on the generated question and computes the geometric ground truth. The model then switches to the solver role and independently derives an answer under the hard constraint of the deterministic ground truth. This design encourages the model to actively probe the boundaries of its own spatial cognition during the question-generation phase, while correcting geometric hallucinations against the absolute ground truth supplied by the environment during the answering phase, forming a continuous self-reinforcement loop within the policy space.

\subsubsection{Task-Adaptive Scheduling}
To realize curriculum learning without manual intervention, SpatialEvo introduces a lightweight task scheduler that dynamically modulates the task sampling distribution. The scheduler first infers the physically realizable feasible task set $\mathcal{T}_s^{\text{feasible}} \subseteq \mathcal{T}$ for the current scene. It then maintains, for each task category $k$, a cumulative score $S_k$ and a sample count $N_k$, and estimates the historical effective accuracy $\bar{a}_k$ via pseudo-observation smoothing to mitigate estimation instability caused by sparse samples in the early stages of training. The sampling weight $w_k$ is negatively correlated with $\bar{a}_k$, and a minimum exploration weight $\delta$ is introduced to prevent task categories that have been thoroughly mastered from being entirely excluded from the sampling pool. This mechanism automatically concentrates training resources on the solver's current cognitive weak spots; as accuracy levels shift across task categories, the training focus migrates accordingly, giving rise to a fully adaptive curriculum driven endogenously by model performance, without any manually prescribed difficulty sequence. The detailed implementation and derivation of the scheduler are provided in Appendix~\ref{app:C.1}.

\subsubsection{Questioner Reward Design}
The Questioner's optimization objective focuses on generating a physically valid, visually grounded question of high quality for a given task category. The Questioner reward comprises two components: format compliance and substantive quality, with $\alpha = 0.1$:
$$r_Q = \alpha\, f_{\text{fmt}} + (1-\alpha)\, f_{\text{valid}} \cdot f_{\text{obs}}.$$
Here $f_{\text{fmt}}$ measures whether the output conforms to predefined structural constraints; $f_{\text{valid}}$ is the geometric validity score from DGE verification; and $f_{\text{obs}}$ is a visual observation quality score that assesses whether the Questioner has sufficiently leveraged the global spatial information present in the multi-view images. $f_{\text{obs}}$ is evaluated by a lightweight LLM judge without relying on a powerful VLM, which scores only the Questioner's textual observation description, examining whether it exhibits a natural perceptual hierarchy from global scene to local target and provides adequate visual grounding for the generated question; scoring criteria are detailed in Appendix~\ref{app:C.2}. The coupled term $f_{\text{valid}} \cdot f_{\text{obs}}$ carries a critical gating semantics: a positive signal is contributed only when a question satisfies both geometric validity and sufficient visual observation simultaneously, preventing the model from generating ``superficially valid'' questions that conform to format but lack genuine spatial understanding. Severe structural errors additionally trigger hard penalties in the implementation, which are omitted here for brevity.

\subsubsection{Solver Reward Design}
The Solver receives reward signals from all candidate questions passed by the Questioner, with both valid and invalid questions serving as effective learning signal. For valid questions, $f_{\text{acc}}$ is anchored directly to the geometric ground truth produced by the DGE. For invalid questions, the DGE outputs a specific invalidation reason alongside its rejection, and the Solver is required to explicitly analyze and explain this reason in light of the task type, question generation rules, and visual observations; explanation quality is scored by a lightweight LLM judge that evaluates whether the model accurately identifies the core failure cause and correctly attributes it to task rules or scene constraints. Detailed evaluation functions for each task category and explanation scoring criteria are provided in Appendix~\ref{app:C.3}. Format compliance is incorporated into the Solver reward via $f_{\text{fmt}}$ to enforce adherence to predefined structural requirements. The unified Solver reward is thus written as, with $\alpha = 0.1$:
$$r_A = \begin{cases} \alpha\, f_{\text{fmt}} + (1-\alpha)\, f_{\text{acc}}, & \text{if } Q \text{ is valid},\\[4pt] \alpha\, f_{\text{fmt}} + (1-\alpha)\, f_{\text{explain}}, & \text{if } Q \text{ is invalid}, \end{cases}$$
where $f_{\text{fmt}}$ measures adherence to the expected output format, $f_{\text{acc}}$ measures agreement with DGE ground truth, and $f_{\text{explain}}$ measures the quality of the explanation for the invalidation reason. This design ensures that invalid questions also serve as meaningful learning signal, further deepening the model's understanding of spatial task rules and geometric constraints. As with the Questioner, severe formatting errors trigger additional hard penalties in the implementation, which are omitted in favor of the weighted main form presented here.

\subsubsection{GRPO Training Procedure}
\label{3.3.5:GRPO}
We adopt the GRPO algorithm to drive the co-evolution of the questioner and solver. For each training scene, the scheduler samples a task type and directs the questioner to generate $n$ candidate questions $\{Q^{(i)}\}_{i=1}^{n}$. The DGE performs legality verification on each $Q^{(i)}$; for questions that pass, it computes the geometric ground truth $G^{(i)}$ and assigns questioner reward $r_Q^{(i)}$; for questions that fail, it returns a specific invalidity reason. The $n$ candidate questions form a single GRPO group for the questioner, within which advantage values are computed independently based on $\{r_Q^{(i)}\}_{i=1}^{n}$.

The $n$ candidate questions are then deduplicated semantically to yield $m$ ($m \leq n$) unique questions $\{Q^{(i)}\}_{i=1}^{m}$, which are passed to the solver. For each unique question $Q^{(i)}$, the solver independently samples $n$ candidate answers $\{A^{(i,j)}\}_{j=1}^{n}$ and computes the solver reward $r_A^{(i,j)}$ against $G^{(i)}$; for invalid questions, the solver generates an explanation based on the invalidity reason returned by the DGE, with the reward assigned by the LLM judge. The solver forms $m$ independent GRPO groups, one per unique question, with advantage values computed independently within each group from the nn
n candidate answers.

Advantage values are computed independently within their respective groups to eliminate reward-scale bias introduced by inherent difficulty variation across scenes:
\begin{equation}
    \hat{A}^{(i)} = \frac{r^{(i)} - \operatorname{mean}\!\left(\{r^{(i)}\}\right)}{\operatorname{std}\!\left(\{r^{(i)}\}\right) + \epsilon}
\end{equation}
The questioner and solver share a single set of model parameters, and gradients from both sides are applied jointly at each training step.

\section{Experiments}
\label{4:Experimental}

\subsection{Experimental Setup}
\label{Experimental Setup}
\paragraph{Environment and Data Configuration.}
We construct the DGE from the training splits of ScanNet~\cite{dai2017scannet}, ScanNet++~\cite{yeshwanth2023scannet++}, and ARKitScenes~\cite{baruch2021arkitscenes}, comprising approximately 4K source scenes in total. The DGE derives noise-free geometric supervision from dense reconstructions and multi-view streams, while requiring only standard RGB images as model input. Dataset and scene statistics are reported in Appendix~\ref{app:D}. We apply SpatialEvo to both Qwen2.5-VL-3B-Instruct~\cite{bai2025qwen25vltechnicalreport} and Qwen2.5-VL-7B-Instruct~\cite{bai2025qwen25vltechnicalreport} as backbone models.

\paragraph{Training Protocol.}
SpatialEvo trains entirely via online reinforcement learning using the GRPO framework, without any supervised fine-tuning stage. The Questioner and Solver share a single policy model and continuously interact with the DGE, enabling self-evolving spatial intelligence through deterministic geometric feedback. Full training details are provided in Appendix~\ref{app:C}.

\paragraph{Baselines.}
We compare against methods spanning the key training paradigms discussed in Section~\ref{intro}. {SpatialLadder}~\citep{li2025spatialladder}, {SpaceR-SFT}~\citep{ouyang2025spacer}, and {ViLaSR}~\citep{wu2025reinforcing} represent static data tuning or RL-based training with either curated annotations or fixed reward functions, without access to deterministic geometric feedback. {Spatial-SSRL}~\citep{liu2025spatial} adopts a self-supervised RL paradigm that derives training signals directly from RGB images, probing whether annotation-free generalization can match geometry-grounded self-evolution.

\paragraph{Evaluation Benchmarks.}
We evaluate across nine benchmarks covering spatial reasoning and general visual understanding. {VSI-Bench}~\cite{yang2025thinking} is the primary metric, assessing quantitative multi-view spatial reasoning including object size, distance, and relative orientation, evaluated at 32 frames. {EmbSpatial}~\cite{du2024embspatial} and {ViewSpatial}~\cite{li2025viewspatial} target embodied spatial understanding and perspective-dependent reasoning. {RealWorldQA}~\cite{zhang2024mme} tests spatial comprehension in open real-world scenes. {V-STAR}~\cite{cheng2025v} benchmarks video spatio-temporal reasoning. {SpatialViz}~\cite{wangspatialviz} evaluates spatial visualization and object transformation reasoning. {STARE}~\cite{li2025unfolding} assesses multi-step visual simulation across geometric transformations and real-world spatial tasks. CoreCognition probes~\cite{li2024core} core cognitive knowledge spanning both spatial and general perceptual reasoning. {MMStar}~\cite{chen2024we} verifies that spatial training does not degrade general visual capabilities. All benchmarks follow their original evaluation protocols unless otherwise noted.

\subsection{Main Results}
\label{Main Results}

\begin{table*}[t]
\centering
\resizebox{\textwidth}{!}{
{\scriptsize
\begin{tabular}{l
 *{3}{>{\centering\arraybackslash}m{1.2cm}}
 >{\centering\arraybackslash}m{1.2cm}
 >{\columncolor{synthcol}\centering\arraybackslash}m{1.2cm}
 *{3}{>{\centering\arraybackslash}m{1.2cm}}
 >{\centering\arraybackslash}m{1.2cm}
 >{\columncolor{synthcol}\centering\arraybackslash}m{1.2cm}
}
\toprule
\multirow{2}{*}{\textbf{Benchmark}} &
\multicolumn{5}{c}{\textbf{Qwen2.5-VL-3B}} &
\multicolumn{5}{c}{\textbf{Qwen2.5-VL-7B}} \\
\cmidrule(lr){2-6} \cmidrule(lr){7-11}
& Baseline & SpatialLadder & SpaceR & SpatialSSRL & \cellcolor{synthcol}SpatialEvo
& Baseline & VILASR & SpaceR & SpatialSSRL & \cellcolor{synthcol}SpatialEvo \\
\midrule
VSI-Bench~\citep{yang2025thinking}      & 28.1 & \textbf{45.7} & 36.0 & 28.0 & 39.2 & 31.1 & 45.4 & 36.8 & 33.7 & \textbf{46.1} \\
RealWorldQA~\citep{zhang2024mme}    & 63.4 & 57.1 & 61.4 & 65.4 & \textbf{66.5} & 69.5 & 57.9 & 64.7 & \textbf{69.9} & 66.7 \\
EmbSpatial~\citep{du2024embspatial}     & 55.9 & 57.6 & 55.6 & 59.8 & \textbf{61.2} & 63.6 & 47.8 & 60.3 & \textbf{69.3} & 66.0 \\
SpatialViz~\citep{wangspatialviz}     & 24.2 & 28.6 & \textbf{31.9} & 25.9 & 25.4 & 27.0 & 29.8 & \textbf{30.9} & 28.4 & 28.6 \\
STARE~\citep{li2025unfolding}          & 33.1 & 26.4 & 36.8 & 36.8 & \textbf{36.9} & 41.8 & 21.4 & 36.2 & \textbf{43.3} & 41.3 \\
CoreCognition~\citep{li2024core}  & 56.8 & \textbf{58.3} & 29.1 & 57.6 & 57.4 & 59.6 & 56.4 & 56.4 & \textbf{60.2} & \textbf{60.2} \\
ViewSpatial~\citep{li2025viewspatial}    & 36.2 & 43.0 & 35.9 & 38.4 & \textbf{42.3} & 36.4 & 32.3 & 35.1 & 37.5 & \textbf{43.2} \\
V-STAR~\citep{cheng2025v}         & 74.9 & 36.7 & 75.4 & \textbf{77.0} & 75.4 & 78.5 & 35.6 & 73.8 & \textbf{79.1} & 78.0 \\
MMStar~\citep{chen2024we}         & 54.6 & 45.8 & 44.9 & \textbf{56.5} & 55.2 & 61.6 & 60.8 & 54.9 & \textbf{63.5} & 62.5 \\
\midrule
\textbf{AVG}            & 47.5 & 44.4 & 45.2 & 49.5 & \textbf{51.1} & 52.1 & 43.0 & 49.9 & 53.9 & \textbf{54.7} \\
\bottomrule
\end{tabular}
}
}
\vspace{7pt}
\caption{\textbf{Main Results.} Nine benchmarks spanning spatial reasoning and general visual understanding. Shaded columns indicate SpatialEvo (ours). \textbf{Bold} denotes the best result per benchmark under each backbone.}
\label{tab:main_results}
\vspace{-15pt}
\end{table*}

Table~\ref{tab:main_results} reports results across nine benchmarks for both Qwen2.5-VL-3B and Qwen2.5-VL-7B backbones. SpatialEvo achieves the highest average score under both settings, reaching 51.1 (3B) and 54.7 (7B), outperforming all baselines by a consistent margin across both model scales.

\paragraph{Core Spatial Reasoning.}
SpatialEvo yields substantial improvements on VSI-Bench, EmbSpatial, and ViewSpatial, which directly correspond to the multi-view scene-level and perspective-dependent task categories covered by our DGE. On VSI-Bench, SpatialEvo achieves 39.2 (3B) and 46.1 (7B), surpassing all baselines including SpatialLadder (45.4) and SpaceR (36.8) under the 7B setting. Consistent gains are similarly observed on EmbSpatial and ViewSpatial, confirming that DGE-driven self-evolution effectively internalizes both embodied and perspective-dependent spatial reasoning and transfers well to held-out benchmarks.

\paragraph{General Capability Retention.}
SpatialEvo maintains competitive performance on MMStar and RealWorldQA, demonstrating that spatial specialization does not degrade general visual understanding. On MMStar, SpatialEvo scores 55.2 (3B) and 62.5 (7B), remaining close to the untuned baseline while annotation-dependent methods such as SpatialLadder (45.8) and ViLaSR (60.8) fall noticeably below. On RealWorldQA, SpatialEvo achieves 66.5 (3B) and 66.7 (7B), outperforming both the baseline and all spatial-specialized competitors under the 3B setting. Spatial-SSRL also retains general capabilities reasonably well; however, its overall average remains below SpatialEvo in both settings, indicating that annotation-free generalization alone is insufficient to match geometry-grounded self-evolution.

\paragraph{Baseline Degradation Analysis.}
The degradation patterns among baselines further contextualize SpatialEvo's advantage. SpatialLadder and ViLaSR collapse on V-STAR to approximately 36, far below the baseline of 74.9 (3B) and 78.5 (7B), indicating that their training pipelines introduce distributional shifts that impair spatio-temporal reasoning. SpaceR similarly drops to 29.1 on CoreCognition (3B, baseline 56.8), suggesting that fixed reward functions tailored to specific spatial tasks can suppress broader cognitive capabilities. SpatialEvo achieves the most favorable balance between spatial specialization and general capability preservation across all evaluated configurations.

\subsection{Ablation Studies}
\label{Ablation Studies}
\begin{table}[t]
    \centering
    \small
    \setlength{\tabcolsep}{4.0pt}
    \resizebox{\columnwidth}{!}{
        \begin{tabular}{lccccccccccr}
        \toprule
        \textbf{Variant} &
        \textbf{VSI-Bench} & \textbf{RealWorldQA} &
        \textbf{EmbSpatial} & \textbf{SpatialViz} & \textbf{STARE} &
        \textbf{CoreCognition} & \textbf{ViewSpatial} &
        \textbf{V-STAR} & \textbf{MMStar} & \textbf{Avg} & \textbf{$\Delta$Avg} \\
        \midrule
        SpatialEvo (Ours) & 46.1 & 66.7 & 66.0 & 28.6 & 41.3 & 60.2 & 43.2 & 78.0 & 62.5 & \textbf{54.7} & -- \\
        \midrule
        \rowcolor{gray!8}
        \multicolumn{12}{l}{\textit{Architecture Design}} \\
        \quad w/o Questioner         & 40.2 & 67.1 & 65.7 & 27.7 & 39.0 & 60.5 & 40.4 & 77.0 & 59.9 & 53.1 & \textcolor{red}{$\downarrow$ 1.6} \\
        \quad w/o Solver             & 36.6 & 70.2 & 61.8 & 26.1 & 39.8 & 58.5 & 34.5 & 75.4 & 60.5 & 51.5 & \textcolor{red}{$\downarrow$ 3.2} \\
        \quad w/o Physical Grounding & 18.8 & 68.5 & 63.5 & 23.4 & 39.7 & 59.7 & 35.4 & 77.0 & 60.6 & 49.6 & \textcolor{red}{$\downarrow$ 5.1} \\
        \quad w/o Adaptive Scheduler & 43.4 & 68.5 & 68.0 & 27.5 & 39.5 & 60.3 & 43.2 & 77.0 & 62.4 & 54.4 & \textcolor{red}{$\downarrow$ 0.3} \\
        \midrule
        \rowcolor{gray!8}
        \multicolumn{12}{l}{\textit{Questioner Reward}} \\
        \quad w/o Validity Reward    & 41.2 & 69.7 & 64.3 & 29.3 & 39.2 & 60.7 & 41.3 & 76.4 & 62.6 & 53.9 & \textcolor{red}{$\downarrow$ 0.8} \\
        \quad w/o Observation Reward & 43.6 & 69.2 & 68.1 & 28.0 & 41.1 & 59.1 & 41.8 & 78.5 & 61.0 & 54.5 & \textcolor{red}{$\downarrow$ 0.2} \\
        \midrule
        \rowcolor{gray!8}
        \multicolumn{12}{l}{\textit{Solver Reward}} \\
        \quad w/o Explanation Reward & 42.9 & 65.1 & 65.9 & 30.8 & 39.5 & 59.9 & 40.9 & 81.7 & 62.3 & 54.3 & \textcolor{red}{$\downarrow$ 0.4} \\
        \bottomrule
        \end{tabular}
    }
    \caption{Ablation study on SpatialEvo (Qwen2.5-VL-7B). $\Delta$Avg denotes performance drop relative to the full model. \textbf{Bold} denotes the best Avg.}
    \label{tab:ablation}
\end{table}
To systematically validate the contribution of each core component in SpatialEvo, we conduct ablation experiments across three design dimensions; results are summarized in Table~\ref{tab:ablation}.

\paragraph{Architectural Design.}
Removing the Questioner (\textit{w/o Questioner}) degrades average performance to 53.1, with notable drops on VSI-Bench (40.2) and ViewSpatial (40.4), as online self-play is replaced by offline data. Removing the Solver (\textit{w/o Solver}) produces a larger drop to 51.5, with VSI-Bench declining sharply to 36.6, confirming that online geometric derivation is essential for internalizing spatial reasoning. Replacing DGE ground truth with majority-voting pseudo-GT (\textit{w/o Physical Grounding}) yields the largest degradation, with the average collapsing to 49.6 and VSI-Bench plummeting to 18.8; the majority-voting mechanism consolidates systematic prediction bias into pseudo-ground-truth, fundamentally corrupting the training signal for geometry-intensive tasks. Additional details are in Appendix~\ref{app:ablation_vote_pseudo_gt}. Removing the adaptive task scheduler (\textit{w/o Adaptive Scheduler}) results in a marginal drop to 54.4, confirming that accuracy-driven dynamic allocation outperforms uniform random sampling.

\paragraph{Questioner Reward.}
Removing the geometric validity reward (\textit{w/o Validity Reward}) degrades performance to 53.9, with VSI-Bench declining to 41.2, as the proportion of physically invalid questions generated by the Questioner increases substantially, reducing the volume of valid questions passed to the Solver and degrading overall training signal quality. Removing the visual observation reward (\textit{w/o Observation Reward}) leads to a smaller but consistent drop to 54.5, as the Questioner bypasses holistic scene perception and generates questions without sufficient visual grounding, resulting in shallower reasoning depth in the Solver.

\paragraph{Solver Reward.}
Removing the explanation reward for invalid questions (\textit{w/o Explanation Reward}) causes performance to decline to 54.3. The effect is most visible on spatially complex benchmarks such as VSI-Bench (42.9 vs.\ 46.1) and ViewSpatial (40.9 vs.\ 43.2), as the geometric constraint information carried by invalid questions is entirely discarded, depriving the Solver of the opportunity to learn task rules from the Questioner's errors and validating the design choice of converting invalid questions into effective learning signal.

\section{Analysis and Discussion}
\label{5:Analysis and Discussion}

\subsection{Online Evolution vs.\ Static Learning}
\label{5.1: Online Evolution}
To evaluate the practical gains of online self-evolution over static data training, we conduct a controlled comparison along two dimensions: reinforcement learning and supervised fine-tuning. We adopt Qwen2.5-VL-3B as the backbone throughout. To align with the data sources of SpatialLadder (26K) and SpaceR (151K), SpatialEvo is restricted to ScanNet scene assets with six core task categories: object counting, object size, room size, absolute distance, relative distance, and relative direction. It is worth noting that SpatialLadder and SpaceR additionally include the \textit{Appearance Order} task, giving them broader task coverage than the SpatialEvo configuration used here. SpatialSSRL (81K) is included in the SFT comparison not for task coverage alignment, but to assess whether its self-supervised training signal generalizes sufficiently to VSI-Bench's spatial task categories.

For the RL comparison, we select SpatialLadder as the reference given its closest task coverage to our configuration, using identical GRPO training settings. For the SFT comparison, samples produced offline during SpatialEvo's online training are curated for supervised fine-tuning, compared against SFT on the three static datasets. SpatialEvo's online training is continued until the number of recorded unique QA pairs reaches approximately 20K. Full training and data details are provided in Appendix~\ref{app:C}.

Results are shown in Table~\ref{tab:paradigm}. Despite operating under a narrower task scope, SpatialEvo surpasses the SpatialLadder RL baseline and outperforms all static dataset SFT counterparts, achieving the highest average of 46.3 and 43.9 respectively. This outcome reveals the core advantage of online self-evolution: the training distribution of a static dataset is frozen at generation time, whereas SpatialEvo continuously aligns the distribution of training samples to the solver's current cognitive frontier through real-time interaction between the Questioner and the DGE, realizing adaptive hard-sample mining that static datasets cannot replicate.

\begin{table}[t]
    \centering
    \small
    \renewcommand{\arraystretch}{1.1}
    \resizebox{\columnwidth}{!}{
    \begin{tabular}{l l c c c c c c c c c}
        \toprule
        \multirow{2}{*}{\textbf{Paradigm}} 
        & \multirow{2}{*}{\textbf{Method}} 
        & \multicolumn{4}{c}{\textbf{Numerical Question}} 
        & \multicolumn{4}{c}{\textbf{Multiple-Choice Question}} 
        & \multirow{2}{*}{\textbf{Avg.}} \\
        \cmidrule(lr){3-6} \cmidrule(lr){7-10}
        & & \textbf{Obj. Count} & \textbf{Abs. Dist.} & \textbf{Obj. Size} & \textbf{Room Size} 
        & \textbf{Rel. Dist.} & \textbf{Rel. Dir.} & \textbf{Route Plan} & \textbf{Appr. Order} \\
        \midrule
        & Qwen2.5-VL-3B (Base) & 33.5 & 21.1 & 17.9 & 22.6 & 32.8 & 42.7 & 29.9 & 21.0 & 28.0 \\
        \midrule
        \multirow{2}{*}{RL}
        & w/ SpatialLadder RL  & 62.8 & 29.8 & 59.2 & 27.8 & 38.9 & 44.2 & \textbf{33.5} & 24.9 & 40.1 \\
        & \cellcolor{gray!10}\textbf{SpatialEvo (Online RL)} & \cellcolor{gray!10}\textbf{65.2} & \cellcolor{gray!10}\textbf{35.1} & \cellcolor{gray!10}\textbf{61.3} & \cellcolor{gray!10}\textbf{51.4} & \cellcolor{gray!10}\textbf{46.2} & \cellcolor{gray!10}\textbf{44.3} & \cellcolor{gray!10}29.9 & \cellcolor{gray!10}\textbf{26.5} & \cellcolor{gray!10}\textbf{46.3} \\
        \midrule
        \multirow{4}{*}{SFT}
        & w/ SpatialLadder Data & \textbf{63.0} & \textbf{31.9} & \textbf{61.2} & 43.0 & \textbf{43.0} & 43.6 & 32.5 & 31.7 & 43.7 \\
        & w/ SpaceR Data        & 28.3 & 25.9 & 36.3 & 34.9 & 35.1 & \textbf{46.8} & \textbf{35.1} & \textbf{43.4} & 36.3 \\
        & w/ SpatialSSRL Data  & 35.6 & 24.6 & 15.4 & 21.8 & 34.1 & 39.2 & 28.9 & 23.5 & 28.1 \\
        & \cellcolor{gray!10}\textbf{w/ SpatialEvo Offline Data} & \cellcolor{gray!10}62.6 & \cellcolor{gray!10}28.9 & \cellcolor{gray!10}60.0 & \cellcolor{gray!10}\textbf{49.1} & \cellcolor{gray!10}40.3 & \cellcolor{gray!10}45.7 & \cellcolor{gray!10}31.4 & \cellcolor{gray!10}25.9 & \cellcolor{gray!10}\textbf{43.9} \\
        \bottomrule
    \end{tabular}
    }
    \caption{Comparison of online self-evolution and static learning paradigms on VSIBench. SpatialEvo outperforms static RL and SFT baselines across both training paradigms.}
    \label{tab:paradigm}
\end{table}

\subsection{Curriculum Emergence Analysis}
\label{5.3: Curriculum Emergence Analysis }

\begin{figure}[htbp]
    \centering
    \begin{minipage}{0.33\textwidth}
        \centering
        \includegraphics[width=\linewidth]{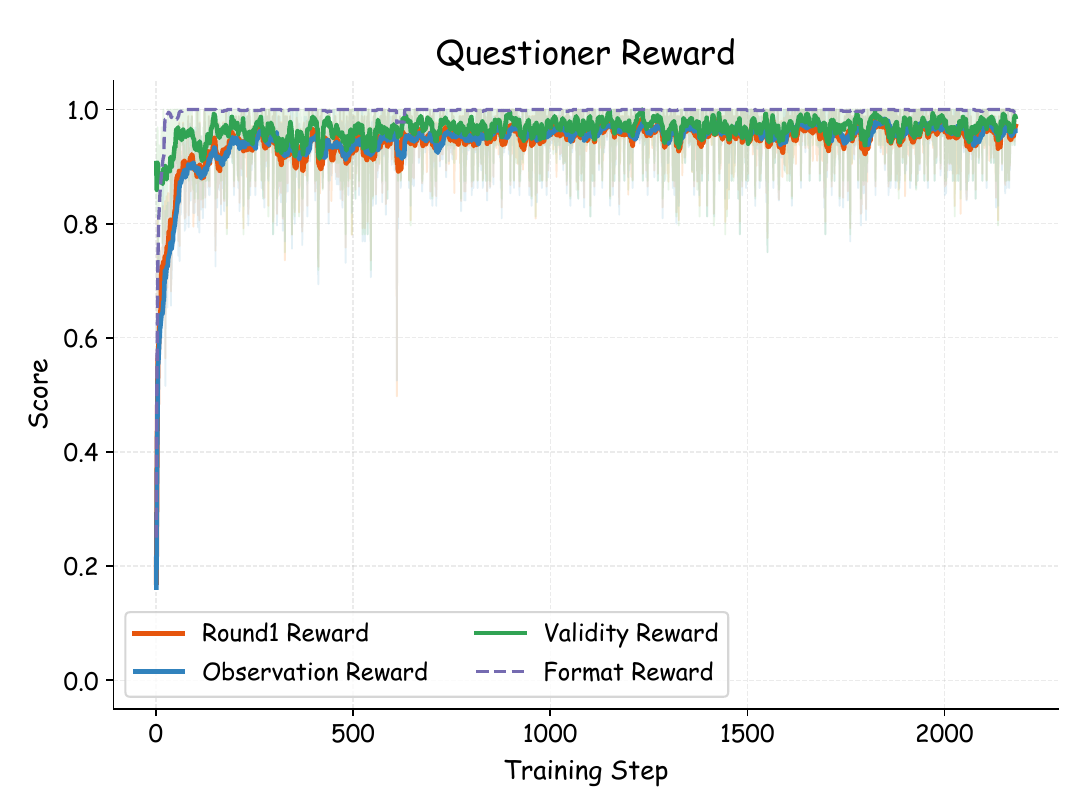}
    \end{minipage}
    \hfill
    \begin{minipage}{0.32\textwidth}
        \centering
        \includegraphics[width=\linewidth]{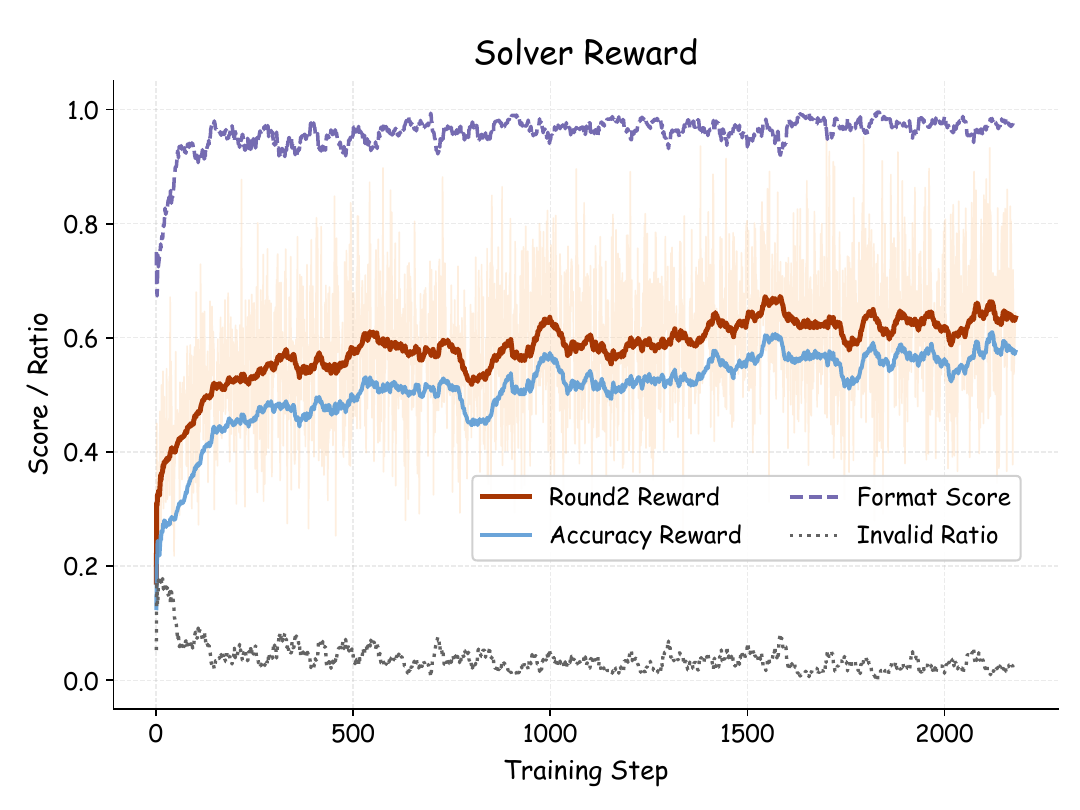}
    \end{minipage}
    \hfill
    \begin{minipage}{0.33\textwidth}
        \centering
        \includegraphics[width=\linewidth]{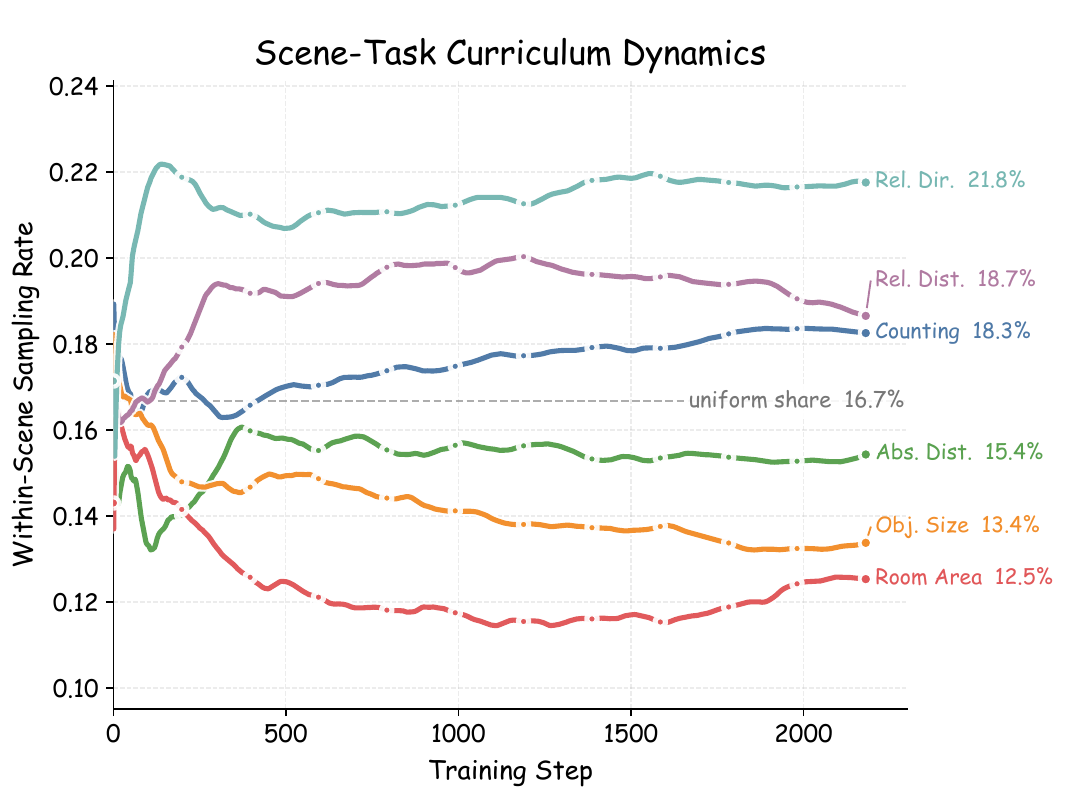}
    \end{minipage}
    \caption{Training dynamics of SpatialEvo. Questioner rewards stabilize as valid question generation is learned (Left); Solver accuracy improves while the invalid ratio declines (Middle); the Adaptive Scheduler up-weights harder categories (Rel.\ Dir., Rel.\ Dist.) and down-weights easier ones (Right).}
    \label{fig:reward}
\end{figure}

\begin{table}[t]
    \centering
    \small
    \renewcommand{\arraystretch}{1.1}
    \resizebox{\columnwidth}{!}{
    \begin{tabular}{l cccc cccc c}
        \toprule
        \multirow{2}{*}{\textbf{Method}} 
        & \multicolumn{4}{c}{\textbf{Numerical Question}} 
        & \multicolumn{4}{c}{\textbf{Multiple-Choice Question}} 
        & \multirow{2}{*}{\textbf{Avg.}} \\
        \cmidrule(lr){2-5} \cmidrule(lr){6-9}
        & \textbf{Obj. Count} & \textbf{Abs. Dist.} & \textbf{Obj. Size} & \textbf{Room Size} 
        & \textbf{Rel. Dist.} & \textbf{Rel. Dir.} & \textbf{Route Plan} & \textbf{Appr. Order} \\
        \midrule
        Qwen2.5-VL-7B (Base)           & 30.7 & 10.2 & 37.14 & 43.40 & 37.2 & 37.1 & 32.0 & 27.8 & 31.1 \\
        \midrule
        \rowcolor{gray!8}
        \multicolumn{10}{l}{\cellcolor{gray!8}\textit{SpatialEvo}} \\
        \quad Iter 1         & 60.6 & 28.9 & 60.3 & 42.5 & 42.7 & 44.4 & \textbf{32.0} & 30.7 & 44.2 \\
        \quad Iter 2         & 63.3 & 28.9 & 62.5 & 33.3 & 43.4 & 43.8 & 30.9 & 36.7 & 45.0 \\
        \quad Iter 3         & \textbf{65.0} & 26.9 & \textbf{63.4} & \textbf{47.3} & 43.0 & 43.4 & 28.9 & 32.9 & 45.1 \\
        \quad Iter 4         & 62.5 & \textbf{32.8} & 61.6 & 35.7 & \textbf{45.1} & \textbf{43.9} & 27.3 & \textbf{40.1} & \textbf{46.1} \\
        \midrule
        \rowcolor{gray!8}
        \multicolumn{10}{l}{\cellcolor{gray!8}\textit{SpatialEvo w/o Adaptive Scheduler}} \\
        \quad Iter 1         & 55.9 & 22.4 & \textbf{61.6} & \textbf{51.8} & \textbf{44.5} & \textbf{43.8} & \textbf{33.0} & \textbf{36.3} & 44.2 \\
        \quad Iter 2         & \textbf{62.1} & 28.6 & 62.0 & 44.0 & 41.0 & 43.3 & 28.9 & 34.3 & \textbf{44.5} \\
        \quad Iter 3         & 57.8 & 29.3 & \textbf{61.6} & 35.8 & 42.3 & 43.4 & 32.5 & 31.9 & 43.7 \\
        \quad Iter 4         & 60.2 & \textbf{30.6} & 58.7 & 39.1 & 40.1 & 43.5 & 28.4 & 32.2 & 43.4 \\
        \bottomrule
    \end{tabular}
    }
    \caption{Effect of the Adaptive Scheduler across iterative self-evolution on VSI-Bench subtasks. Both groups are initialized from the same base model and trained for four iterations; the only difference is whether the Adaptive Scheduler is enabled.}
    \label{tab:adaptive_scheduler}
\end{table}

\paragraph{Effect of the Adaptive Scheduler across iterations.}
Table~\ref{tab:adaptive_scheduler} compares SpatialEvo with and without the Adaptive Scheduler across four iterative self-evolution rounds on VSI-Bench. Without the scheduler, the model performs comparably in early iterations (44.2 at Iter 1, 44.5 at Iter 2) but subsequently stagnates and declines, reaching only 43.4 at Iter 4, as uniform task sampling fails to concentrate training resources on categories where the model remains weak. In contrast, the full SpatialEvo with the Adaptive Scheduler exhibits monotonically increasing average performance across all four iterations (44.2 $\to$ 45.0 $\to$ 45.1 $\to$ 46.1), with particularly strong late-stage gains on Abs.\ Dist.\ (32.8), Rel.\ Dist.\ (45.1), and Appr.\ Order (40.1) at Iter 4, categories that receive increasing sampling weight as the scheduler identifies them as persistent weak spots.

\paragraph{Reward trajectories and curriculum emergence.}
As shown in Figure~\ref{fig:reward}, both Questioner and Solver rewards improve steadily throughout training. The Validity Reward rises rapidly and stabilizes near 1.0, indicating that the Questioner quickly learns to generate physically valid questions; the Accuracy Reward of the Solver increases consistently while the Invalid Ratio declines, confirming progressive internalization of geometric reasoning. The Scene-Task Curriculum Dynamics plot reveals how this improvement is shaped: sampling rates diverge from the uniform share (16.7\%) as training progresses, with harder categories such as Rel.\ Dir.\ (21.8\%) and Rel.\ Dist.\ (18.7\%) being up-weighted and easier categories such as Room Area (12.5\%) and Obj.\ Size (13.4\%) being down-weighted. This endogenously emergent curriculum, driven entirely by historical accuracy without manual design, explains the sustained performance improvement observed in Table~\ref{tab:adaptive_scheduler}.

\section{Conclusions}
\label{Conclusions}
We present SpatialEvo, the first framework to introduce the self-evolving paradigm into 3D spatial reasoning. Unlike conventional self-evolving methods that rely on model voting to construct pseudo-labels, SpatialEvo exploits the unique property of spatial reasoning that visual inputs inherently carry physical information such as 3D point clouds and camera poses, enabling programmatic computation of exact ground truth. This transforms unannotated 3D scene assets into zero-noise reward judges, replacing model consensus with deterministic physical feedback. A single policy model co-evolves as Questioner and Solver under DGE constraints, with a task scheduler enabling adaptive curriculum self-emergence. Experiments validate significant gains across multiple spatial reasoning benchmarks.

We hope the physically grounded self-evolution paradigm explored by SpatialEvo can serve as a reference for broader embodied intelligence research. When self-exploration is rooted in a verifiable physical environment, the continual emergence of spatial reasoning capability is driven endogenously by the model's interaction with the objective world, rather than costly human annotation.

\bibliographystyle{plainnat}
\bibliography{main}

@inproceedings{chen2024spatialvlm,
  title={Spatialvlm: Endowing vision-language models with spatial reasoning capabilities},
  author={Chen, Boyuan and Xu, Zhuo and Kirmani, Sean and Ichter, Brain and Sadigh, Dorsa and Guibas, Leonidas and Xia, Fei},
  booktitle={Proceedings of the IEEE/CVF Conference on Computer Vision and Pattern Recognition},
  pages={14455--14465},
  year={2024}
}

@article{cheng2024spatialrgpt,
  title={Spatialrgpt: Grounded spatial reasoning in vision-language models},
  author={Cheng, An-Chieh and Yin, Hongxu and Fu, Yang and Guo, Qiushan and Yang, Ruihan and Kautz, Jan and Wang, Xiaolong and Liu, Sifei},
  journal={Advances in Neural Information Processing Systems},
  volume={37},
  pages={135062--135093},
  year={2024}
}

@article{li2025embodied,
  title={Embodied intelligence for 3d understanding: A survey on 3d scene question answering},
  author={Li, Zechuan and Yu, Hongshan and Ding, Yihao and Li, Yan and He, Yong and Akhtar, Naveed},
  journal={Information Fusion},
  pages={103624},
  year={2025},
  publisher={Elsevier}
}

@article{ma2024llms,
  title={When llms step into the 3d world: A survey and meta-analysis of 3d tasks via multi-modal large language models},
  author={Ma, Xianzheng and Smart, Brandon and Bhalgat, Yash and Chen, Shuai and Li, Xinghui and Ding, Jian and Gu, Jindong and Chen, Dave Zhenyu and Peng, Songyou and Bian, Jia-Wang and others},
  journal={arXiv preprint arXiv:2405.10255},
  year={2024}
}

@inproceedings{song2025robospatial,
  title={Robospatial: Teaching spatial understanding to 2d and 3d vision-language models for robotics},
  author={Song, Chan Hee and Blukis, Valts and Tremblay, Jonathan and Tyree, Stephen and Su, Yu and Birchfield, Stan},
  booktitle={Proceedings of the Computer Vision and Pattern Recognition Conference},
  pages={15768--15780},
  year={2025}
}

@article{wu2024continual,
  title={Continual learning for large language models: A survey},
  author={Wu, Tongtong and Luo, Linhao and Li, Yuan-Fang and Pan, Shirui and Vu, Thuy-Trang and Haffari, Gholamreza},
  journal={arXiv preprint arXiv:2402.01364},
  year={2024}
}

@article{shi2025continual,
  title={Continual learning of large language models: A comprehensive survey},
  author={Shi, Haizhou and Xu, Zihao and Wang, Hengyi and Qin, Weiyi and Wang, Wenyuan and Wang, Yibin and Wang, Zifeng and Ebrahimi, Sayna and Wang, Hao},
  journal={ACM Computing Surveys},
  volume={58},
  number={5},
  pages={1--42},
  year={2025},
  publisher={ACM New York, NY}
}

@inproceedings{wang2025language,
  title={Language models as continuous self-evolving data engineers},
  author={Wang, Peidong and Wang, Ming and Ma, Zhiming and Yang, Xiaocui and Feng, Shi and Wang, Daling and Zhang, Yifei and Song, Kaisong},
  booktitle={Proceedings of the 2025 Conference on Empirical Methods in Natural Language Processing},
  pages={18108--18127},
  year={2025}
}

@article{he2025visplay,
  title={Visplay: Self-evolving vision-language models from images},
  author={He, Yicheng and Huang, Chengsong and Li, Zongxia and Huang, Jiaxin and Yang, Yonghui},
  journal={arXiv preprint arXiv:2511.15661},
  year={2025}
}

@article{liu2024diving,
  title={Diving into self-evolving training for multimodal reasoning},
  author={Liu, Wei and Li, Junlong and Zhang, Xiwen and Zhou, Fan and Cheng, Yu and He, Junxian},
  journal={arXiv preprint arXiv:2412.17451},
  year={2024}
}

@article{deng2024enhancing,
  title={Enhancing large vision language models with self-training on image comprehension},
  author={Deng, Yihe and Lu, Pan and Yin, Fan and Hu, Ziniu and Shen, Sheng and Gu, Quanquan and Zou, James and Chang, Kai-Wei and Wang, Wei},
  journal={Advances in Neural Information Processing Systems},
  volume={37},
  pages={131369--131397},
  year={2024}
}

@article{zhao2025absolute,
  title={Absolute zero: Reinforced self-play reasoning with zero data},
  author={Zhao, Andrew and Wu, Yiran and Yue, Yang and Wu, Tong and Xu, Quentin and Lin, Matthieu and Wang, Shenzhi and Wu, Qingyun and Zheng, Zilong and Huang, Gao},
  journal={arXiv preprint arXiv:2505.03335},
  year={2025}
}

@article{tao2024survey,
  title={A survey on self-evolution of large language models},
  author={Tao, Zhengwei and Lin, Ting-En and Chen, Xiancai and Li, Hangyu and Wu, Yuchuan and Li, Yongbin and Jin, Zhi and Huang, Fei and Tao, Dacheng and Zhou, Jingren},
  journal={arXiv preprint arXiv:2404.14387},
  year={2024}
}

@article{wu2025reinforcing,
  title={Reinforcing spatial reasoning in vision-language models with interwoven thinking and visual drawing},
  author={Wu, Junfei and Guan, Jian and Feng, Kaituo and Liu, Qiang and Wu, Shu and Wang, Liang and Wu, Wei and Tan, Tieniu},
  journal={arXiv preprint arXiv:2506.09965},
  year={2025}
}

@inproceedings{yang2025thinking,
  title={Thinking in space: How multimodal large language models see, remember, and recall spaces},
  author={Yang, Jihan and Yang, Shusheng and Gupta, Anjali W and Han, Rilyn and Fei-Fei, Li and Xie, Saining},
  booktitle={Proceedings of the Computer Vision and Pattern Recognition Conference},
  pages={10632--10643},
  year={2025}
}

@inproceedings{deng2025self,
  title={Self-Improvement in Multimodal Large Language Models: A Survey},
  author={Deng, Shijian and Wang, Kai and Yang, Tianyu and Singh, Harsh and Tian, Yapeng},
  booktitle={Findings of the Association for Computational Linguistics: EMNLP 2025},
  pages={1987--2006},
  year={2025}
}

@article{li2025self,
  title={Self-rewarding vision-language model via reasoning decomposition},
  author={Li, Zongxia and Yu, Wenhao and Huang, Chengsong and Liu, Rui and Liang, Zhenwen and Liu, Fuxiao and Che, Jingxi and Yu, Dian and Boyd-Graber, Jordan and Mi, Haitao and others},
  journal={arXiv preprint arXiv:2508.19652},
  year={2025}
}

@article{zhai2024fine,
  title={Fine-tuning large vision-language models as decision-making agents via reinforcement learning},
  author={Zhai, Yuexiang and Bai, Hao and Lin, Zipeng and Pan, Jiayi and Tong, Shengbang and Zhou, Yifei and Suhr, Alane and Xie, Saining and LeCun, Yann and Ma, Yi and others},
  journal={Advances in neural information processing systems},
  volume={37},
  pages={110935--110971},
  year={2024}
}

@article{shao2024deepseekmath,
  title={Deepseekmath: Pushing the limits of mathematical reasoning in open language models},
  author={Shao, Zhihong and Wang, Peiyi and Zhu, Qihao and Xu, Runxin and Song, Junxiao and Bi, Xiao and Zhang, Haowei and Zhang, Mingchuan and Li, YK and Wu, Yang and others},
  journal={arXiv preprint arXiv:2402.03300},
  year={2024}
}

@article{shi2025efficient,
  title={Efficient reinforcement finetuning via adaptive curriculum learning},
  author={Shi, Taiwei and Wu, Yiyang and Song, Linxin and Zhou, Tianyi and Zhao, Jieyu},
  journal={arXiv preprint arXiv:2504.05520},
  year={2025}
}

@inproceedings{du2024embspatial,
  title={Embspatial-bench: Benchmarking spatial understanding for embodied tasks with large vision-language models},
  author={Du, Mengfei and Wu, Binhao and Li, Zejun and Huang, Xuan-Jing and Wei, Zhongyu},
  booktitle={Proceedings of the 62nd Annual Meeting of the Association for Computational Linguistics (Volume 2: Short Papers)},
  pages={346--355},
  year={2024}
}

@article{li2025viewspatial,
  title={Viewspatial-bench: Evaluating multi-perspective spatial localization in vision-language models},
  author={Li, Dingming and Li, Hongxing and Wang, Zixuan and Yan, Yuchen and Zhang, Hang and Chen, Siqi and Hou, Guiyang and Jiang, Shengpei and Zhang, Wenqi and Shen, Yongliang and others},
  journal={arXiv preprint arXiv:2505.21500},
  year={2025}
}

@article{li2026mm,
  title={Mm-zero: Self-evolving multi-model vision language models from zero data},
  author={Li, Zongxia and Du, Hongyang and Huang, Chengsong and Wu, Xiyang and Yu, Lantao and He, Yicheng and Xie, Jing and Wu, Xiaomin and Liu, Zhichao and Zhang, Jiarui and others},
  journal={arXiv preprint arXiv:2603.09206},
  year={2026}
}

@inproceedings{cai2025spatialbot,
  title={Spatialbot: Precise spatial understanding with vision language models},
  author={Cai, Wenxiao and Ponomarenko, Iaroslav and Yuan, Jianhao and Li, Xiaoqi and Yang, Wankou and Dong, Hao and Zhao, Bo},
  booktitle={2025 IEEE International Conference on Robotics and Automation (ICRA)},
  pages={9490--9498},
  year={2025},
  organization={IEEE}
}

@article{ma2024spatialpin,
  title={Spatialpin: Enhancing spatial reasoning capabilities of vision-language models through prompting and interacting 3d priors},
  author={Ma, Chenyang and Lu, Kai and Cheng, Ta-Ying and Trigoni, Niki and Markham, Andrew},
  journal={Advances in neural information processing systems},
  volume={37},
  pages={68803--68832},
  year={2024}
}

@article{li2025spatialladder,
  title={Spatialladder: Progressive training for spatial reasoning in vision-language models},
  author={Li, Hongxing and Li, Dingming and Wang, Zixuan and Yan, Yuchen and Wu, Hang and Zhang, Wenqi and Shen, Yongliang and Lu, Weiming and Xiao, Jun and Zhuang, Yueting},
  journal={arXiv preprint arXiv:2510.08531},
  year={2025}
}

@article{thawakar2025evolmm,
  title={Evolmm: Self-evolving large multimodal models with continuous rewards},
  author={Thawakar, Omkar and Venkatraman, Shravan and Thawkar, Ritesh and Shaker, Abdelrahman and Cholakkal, Hisham and Anwer, Rao Muhammad and Khan, Salman and Khan, Fahad},
  journal={arXiv preprint arXiv:2511.16672},
  year={2025}
}

@article{wang2026v,
  title={V-Zero: Self-Improving Multimodal Reasoning with Zero Annotation},
  author={Wang, Han and Yang, Yi and Hu, Jingyuan and Zhu, Minfeng and Chen, Wei},
  journal={arXiv preprint arXiv:2601.10094},
  year={2026}
}

@article{wang2025vision,
  title={Vision-zero: Scalable vlm self-improvement via strategic gamified self-play},
  author={Wang, Qinsi and Liu, Bo and Zhou, Tianyi and Shi, Jing and Lin, Yueqian and Chen, Yiran and Li, Hai Helen and Wan, Kun and Zhao, Wentian},
  journal={arXiv preprint arXiv:2509.25541},
  year={2025}
}

@article{yang2025visual,
  title={Visual spatial tuning},
  author={Yang, Rui and Zhu, Ziyu and Li, Yanwei and Huang, Jingjia and Yan, Shen and Zhou, Siyuan and Liu, Zhe and Li, Xiangtai and Li, Shuangye and Wang, Wenqian and others},
  journal={arXiv preprint arXiv:2511.05491},
  year={2025}
}

@article{cai2025scaling,
  title={Scaling spatial intelligence with multimodal foundation models},
  author={Cai, Zhongang and Wang, Ruisi and Gu, Chenyang and Pu, Fanyi and Xu, Junxiang and Wang, Yubo and Yin, Wanqi and Yang, Zhitao and Wei, Chen and Sun, Qingping and others},
  journal={arXiv preprint arXiv:2511.13719},
  year={2025}
}

@article{ouyang2025spacer,
  title={Spacer: Reinforcing mllms in video spatial reasoning},
  author={Ouyang, Kun and Liu, Yuanxin and Wu, Haoning and Liu, Yi and Zhou, Hao and Zhou, Jie and Meng, Fandong and Sun, Xu},
  journal={arXiv preprint arXiv:2504.01805},
  year={2025}
}

@article{yang2025mmsi,
  title={Mmsi-bench: A benchmark for multi-image spatial intelligence},
  author={Yang, Sihan and Xu, Runsen and Xie, Yiman and Yang, Sizhe and Li, Mo and Lin, Jingli and Zhu, Chenming and Chen, Xiaochen and Duan, Haodong and Yue, Xiangyu and others},
  journal={arXiv preprint arXiv:2505.23764},
  year={2025}
}

@misc{bai2025qwen25vltechnicalreport,
      title={Qwen2.5-VL Technical Report}, 
      author={Shuai Bai and Keqin Chen and Xuejing Liu and Jialin Wang and Wenbin Ge and Sibo Song and Kai Dang and Peng Wang and Shijie Wang and Jun Tang and Humen Zhong and Yuanzhi Zhu and Mingkun Yang and Zhaohai Li and Jianqiang Wan and Pengfei Wang and Wei Ding and Zheren Fu and Yiheng Xu and Jiabo Ye and Xi Zhang and Tianbao Xie and Zesen Cheng and Hang Zhang and Zhibo Yang and Haiyang Xu and Junyang Lin},
      year={2025},
      eprint={2502.13923},
      archivePrefix={arXiv},
      primaryClass={cs.CV},
      url={https://arxiv.org/abs/2502.13923}, 
}

@inproceedings{dai2017scannet,
  title={Scannet: Richly-annotated 3d reconstructions of indoor scenes},
  author={Dai, Angela and Chang, Angel X and Savva, Manolis and Halber, Maciej and Funkhouser, Thomas and Nie{\ss}ner, Matthias},
  booktitle={Proceedings of the IEEE conference on computer vision and pattern recognition},
  pages={5828--5839},
  year={2017}
}

@inproceedings{yeshwanth2023scannet++,
  title={Scannet++: A high-fidelity dataset of 3d indoor scenes},
  author={Yeshwanth, Chandan and Liu, Yueh-Cheng and Nie{\ss}ner, Matthias and Dai, Angela},
  booktitle={Proceedings of the IEEE/CVF International Conference on Computer Vision},
  pages={12--22},
  year={2023}
}

@article{baruch2021arkitscenes,
  title={Arkitscenes: A diverse real-world dataset for 3d indoor scene understanding using mobile rgb-d data},
  author={Baruch, Gilad and Chen, Zhuoyuan and Dehghan, Afshin and Dimry, Tal and Feigin, Yuri and Fu, Peter and Gebauer, Thomas and Joffe, Brandon and Kurz, Daniel and Schwartz, Arik and others},
  journal={arXiv preprint arXiv:2111.08897},
  year={2021}
}

@article{liu2025spatial,
  title={Spatial-ssrl: Enhancing spatial understanding via self-supervised reinforcement learning},
  author={Liu, Yuhong and Zhang, Beichen and Zang, Yuhang and Cao, Yuhang and Xing, Long and Dong, Xiaoyi and Duan, Haodong and Lin, Dahua and Wang, Jiaqi},
  journal={arXiv preprint arXiv:2510.27606},
  year={2025}
}

@article{zhang2024mme,
  title={Mme-realworld: Could your multimodal llm challenge high-resolution real-world scenarios that are difficult for humans?},
  author={Zhang, Yi-Fan and Zhang, Huanyu and Tian, Haochen and Fu, Chaoyou and Zhang, Shuangqing and Wu, Junfei and Li, Feng and Wang, Kun and Wen, Qingsong and Zhang, Zhang and others},
  journal={arXiv preprint arXiv:2408.13257},
  year={2024}
}

@article{cheng2025v,
  title={V-star: Benchmarking video-llms on video spatio-temporal reasoning},
  author={Cheng, Zixu and Hu, Jian and Liu, Ziquan and Si, Chenyang and Li, Wei and Gong, Shaogang},
  journal={arXiv preprint arXiv:2503.11495},
  year={2025}
}

@inproceedings{wangspatialviz,
  title={SpatialViz-Bench: A Cognitively-Grounded Benchmark for Diagnosing Spatial Visualization in MLLMs},
  author={Wang, Siting and Pei, Minnan and Sun, Luoyang and Deng, Cheng and Li, Yuchen and Shao, Kun and Tian, Zheng and Zhang, Haifeng and Wang, Jun},
  booktitle={The Fourteenth International Conference on Learning Representations}
}

@article{li2024core,
  title={Core knowledge deficits in multi-modal language models},
  author={Li, Yijiang and Gao, Qingying and Zhao, Tianwei and Wang, Bingyang and Sun, Haoran and Lyu, Haiyun and Hawkins, Robert D and Vasconcelos, Nuno and Golan, Tal and Luo, Dezhi and others},
  journal={arXiv preprint arXiv:2410.10855},
  year={2024}
}

@article{chen2024we,
  title={Are we on the right way for evaluating large vision-language models?},
  author={Chen, Lin and Li, Jinsong and Dong, Xiaoyi and Zhang, Pan and Zang, Yuhang and Chen, Zehui and Duan, Haodong and Wang, Jiaqi and Qiao, Yu and Lin, Dahua and others},
  journal={Advances in Neural Information Processing Systems},
  volume={37},
  pages={27056--27087},
  year={2024}
}

@article{li2025unfolding,
  title={Unfolding spatial cognition: Evaluating multimodal models on visual simulations},
  author={Li, Linjie and Bigverdi, Mahtab and Gu, Jiawei and Ma, Zixian and Yang, Yinuo and Li, Ziang and Choi, Yejin and Krishna, Ranjay},
  journal={arXiv preprint arXiv:2506.04633},
  year={2025}
}

@article{agarwal2025gpt,
  title={gpt-oss-120b \& gpt-oss-20b model card},
  author={Agarwal, Sandhini and Ahmad, Lama and Ai, Jason and Altman, Sam and Applebaum, Andy and Arbus, Edwin and Arora, Rahul K and Bai, Yu and Baker, Bowen and Bao, Haiming and others},
  journal={arXiv preprint arXiv:2508.10925},
  year={2025}
}

@article{chen2025self,
  title={Self-evolving curriculum for llm reasoning},
  author={Chen, Xiaoyin and Lu, Jiarui and Kim, Minsu and Zhang, Dinghuai and Tang, Jian and Pich{\'e}, Alexandre and Gontier, Nicolas and Bengio, Yoshua and Kamalloo, Ehsan},
  journal={arXiv preprint arXiv:2505.14970},
  year={2025}
}

@article{luo2025geogrambench,
  title={Geogrambench: Benchmarking the geometric program reasoning in modern llms},
  author={Luo, Shixian and Zhu, Zezhou and Yuan, Yu and Yang, Yuncheng and Shan, Lianlei and Wu, Yong},
  journal={arXiv preprint arXiv:2505.17653},
  year={2025}
}

@inproceedings{ma20253dsrbench,
  title={3dsrbench: A comprehensive 3d spatial reasoning benchmark},
  author={Ma, Wufei and Chen, Haoyu and Zhang, Guofeng and Chou, Yu-Cheng and Chen, Jieneng and de Melo, Celso and Yuille, Alan},
  booktitle={Proceedings of the IEEE/CVF International Conference on Computer Vision},
  pages={6924--6934},
  year={2025}
}

@article{xu2025spatialbench,
  title={Spatialbench: Benchmarking multimodal large language models for spatial cognition},
  author={Xu, Peiran and Wang, Sudong and Zhu, Yao and Li, Jianing and Zhang, Yunjian},
  journal={arXiv preprint arXiv:2511.21471},
  year={2025}
}

@inproceedings{mayer2025ivispar,
  title={iVISPAR—An Interactive Visual-Spatial Reasoning Benchmark for VLMs},
  author={Mayer, Julius and Ballout, Mohamad and Jassim, Serwan and Nezami, Farbod Nosrat and Bruni, Elia},
  booktitle={Proceedings of the 2025 Conference on Empirical Methods in Natural Language Processing},
  pages={26745--26769},
  year={2025}
}

@inproceedings{zhou2025vlm4d,
  title={Vlm4d: Towards spatiotemporal awareness in vision language models},
  author={Zhou, Shijie and Vilesov, Alexander and He, Xuehai and Wan, Ziyu and Zhang, Shuwang and Nagachandra, Aditya and Chang, Di and Chen, Dongdong and Wang, Xin Eric and Kadambi, Achuta},
  booktitle={Proceedings of the IEEE/CVF international conference on computer vision},
  pages={8600--8612},
  year={2025}
}

@article{liu2025ssr,
  title={Ssr: Enhancing depth perception in vision-language models via rationale-guided spatial reasoning},
  author={Liu, Yang and Ma, Ming and Yu, Xiaomin and Ding, Pengxiang and Zhao, Han and Sun, Mingyang and Huang, Siteng and Wang, Donglin},
  journal={arXiv preprint arXiv:2505.12448},
  year={2025}
}

@article{li2025mixture,
  title={Mixture-of-Visual-Thoughts: Exploring Context-Adaptive Reasoning Mode Selection for General Visual Reasoning},
  author={Li, Zejun and Zhao, Yingxiu and Zhang, Jiwen and Wang, Siyuan and Yao, Yang and Zhao, Runzhou and Song, Jun and Zheng, Bo and Wei, Zhongyu},
  journal={arXiv preprint arXiv:2509.22746},
  year={2025}
}

@article{zheng2026pearl,
  title={PEARL: Personalized Streaming Video Understanding Model},
  author={Zheng, Yuanhong and An, Ruichuan and Lin, Xiaopeng and Liu, Yuxing and Yang, Sihan and Zhang, Huanyu and Li, Haodong and Zhang, Qintong and Zhang, Renrui and Li, Guopeng and others},
  journal={arXiv preprint arXiv:2603.20422},
  year={2026}
}

@inproceedings{tian2025nuscenes,
  title={Nuscenes-spatialqa: A spatial understanding and reasoning benchmark for vision-language models in autonomous driving},
  author={Tian, Kexin and Mao, Jingrui and Zhang, Yunlong and Jiang, Jiwan and Zhou, Yang and Tu, Zhengzhong},
  booktitle={Proceedings of the IEEE/CVF International Conference on Computer Vision},
  pages={4567--4576},
  year={2025}
}

@article{lyu2024mmscan,
  title={Mmscan: A multi-modal 3d scene dataset with hierarchical grounded language annotations},
  author={Lyu, Ruiyuan and Lin, Jingli and Wang, Tai and Yang, Shuai and Mao, Xiaohan and Chen, Yilun and Xu, Runsen and Huang, Haifeng and Zhu, Chenming and Lin, Dahua and others},
  journal={Advances in Neural Information Processing Systems},
  volume={37},
  pages={50898--50924},
  year={2024}
}

@inproceedings{wang2025spatial457,
  title={Spatial457: A diagnostic benchmark for 6d spatial reasoning of large mutimodal models},
  author={Wang, Xingrui and Ma, Wufei and Zhang, Tiezheng and de Melo, Celso M and Chen, Jieneng and Yuille, Alan},
  booktitle={Proceedings of the Computer Vision and Pattern Recognition Conference},
  pages={24669--24679},
  year={2025}
}
\clearpage

\appendix

\section{Limitations}
\label{app:A}

Although SpatialEvo demonstrates substantial improvements in spatial reasoning, several limitations merit explicit discussion.

\paragraph{Dependency on high-fidelity 3D assets.}
The framework's core reliance on the Deterministic Geometric Environment (DGE) confines its applicability to scenes equipped with complete 3D assets. Specifically, SpatialEvo requires high-quality indoor point cloud reconstructions, calibrated camera pose parameters, and comprehensive scene coverage, which currently restricts its use to static indoor environments such as those in the ScanNet dataset family. In outdoor or dynamic settings, geometric consistency is difficult to guarantee due to sparse point clouds, complex scale variation, or moving objects, thereby undermining the reliability of ground-truth computation.

\paragraph{Sensitivity to entity parsing quality.}
The question parsing stage of the DGE pipeline relies on a language model to extract structured entities from free-form natural language. When questions contain ambiguous references or underspecified targets, parsing errors may arise and propagate into subsequent verification and computation stages, introducing a source of noise that deterministic geometric reasoning alone cannot fully mitigate.

\paragraph{Sensitivity to point cloud quality.}
The DGE's geometric ground-truth computation is inherently sensitive to the fidelity of the underlying point clouds. Reconstruction artifacts, point sparsity, and occlusions can degrade the precision of geometric operators such as bounding box fitting and depth estimation, leading to approximation errors in continuous-valued tasks (e.g., absolute distance and object size estimation). Although relative-error tolerance bands are introduced at the reward stage to partially absorb reconstruction noise, their effectiveness remains fundamentally bounded by the quality of the underlying data.

These limitations largely stem from the reliance on explicit 3D representations. Future work may explore reducing dependence on point clouds via alternative or implicit spatial representations, as well as on-demand geometry construction, to improve scalability and generalization.

\section{Implementation Details of the DGE}
\label{app:B}

This appendix provides a more granular formalization of the methodology introduced in Section~\ref{Deterministic Geometric Environment}, with the aim of enabling full reproducibility of the DGE's decision logic, task scheduling mechanism, and reward definitions. We proceed as follows: complete task-specific validity rules (Appendix~\ref{app:B.1}), and the automated verification pipeline and entity extraction prompts (Appendix~\ref{app:B.2}).

\subsection{Complete Task-Specific Validation Rules}
\label{app:B.1}

\subsubsection{Unified Validation Decomposition}
\label{app:B.1.1}

For any question $Q$, task type $t$, and scene context $x$, the DGE acceptance condition is expressed as
\begin{equation}
\mathrm{Valid}(Q,t,x)
=
\mathbb{I}\!\left[
\mathcal{C}_{\mathrm{mode}}
\wedge
\mathcal{C}_{\mathrm{extract}}
\wedge
\mathcal{C}_{\mathrm{pool}}
\wedge
\mathcal{C}_{\mathrm{schema}}
\wedge
\mathcal{C}_{\mathrm{solver}}
\right],
\end{equation}
where the five conditions are defined as follows. $\mathcal{C}_{\mathrm{mode}}$ requires that the input modality be consistent with the task type, i.e., the input contract for scene-level, single-image, or image-pair inputs is satisfied. $\mathcal{C}_{\mathrm{extract}}$ requires that structured entity extraction succeed with all task-required fields populated. $\mathcal{C}_{\mathrm{pool}}$ requires that all extracted labels fall within the grounded candidate pool that is both observable for the current sample and permitted by the task definition. $\mathcal{C}_{\mathrm{schema}}$ requires that task-level structural constraints be satisfied, including non-empty candidate lists, no repetition across distinct roles, and the query target not appearing within its own candidate set. $\mathcal{C}_{\mathrm{solver}}$ requires that the downstream deterministic geometry operator be executable, meaning all required metadata, camera parameters, object anchors, or detection results are available.

For conciseness, we define the following grounded label pools:
\begin{equation}
\mathcal{U}_{\mathrm{scene}}
=
\{\ell \mid n_{\mathrm{scene}}(\ell)=1\},
\qquad
\mathcal{C}_{\mathrm{scene}}
=
\{\ell \mid n_{\mathrm{scene}}(\ell)\ge 2\},
\end{equation}
where $n_{\mathrm{scene}}(\ell)$ is the instance count of scene-level object label $\ell$. For single-image and image-pair tasks, we define over the per-frame visible object set:
\begin{equation}
\mathcal{U}_{f}(v_{\min})
=
\{\ell \mid n_f(\ell;\, v_{\min}) = 1\},
\end{equation}
where $n_f(\ell;\, v_{\min})$ denotes the number of visible instances of label $\ell$ in frame $f$ under visibility threshold $v_{\min}$. All frame-level visibility checks satisfy $v_{\min}\ge 0.1$; when a specific tool supplies a stricter threshold, the larger value is applied.

In addition to the task-specific rules tabulated below, the DGE uniformly enforces three categories of structural constraints: (i)~\textbf{same-label constraints}, wherein measurement pairs or reference-target pairs must not refer to the same object; (ii)~\textbf{distinct-field-group constraints}, wherein entities assigned to different roles in multi-role orientation tasks must be mutually distinct; and (iii)~\textbf{list constraints}, wherein candidate lists must be non-empty, deduplicated, and must not contain the query target itself.

\subsubsection{Comprehensive Task Introduction and Validation Rules}

Validation rules and detailed descriptions for scene-level spatial reasoning tasks are provided in a unified manner. Each task is characterized by its semantic objective, representative question form, grounding scope, and core validity constraints to ensure well-posedness, while the deterministic solver specifies how ground-truth answers are computed within the DGE framework for consistent and verifiable evaluation. Detailed definitions and examples are summarized in Tables~\ref{tab:scene_tasks}, \ref{tab:single_image_tasks}, and \ref{tab:image_pair_tasks}.

\begin{table}[h]
\centering
\caption{Validation rules and detailed descriptions for scene-level spatial reasoning tasks.}
\label{tab:scene_tasks}
\small
\begin{adjustbox}{max width=\textwidth}
\begin{tabular}{L{2.6cm} L{3.0cm} L{4.2cm} L{1.8cm} L{3.0cm} L{2.5cm}}
\toprule
\textbf{Task} & \textbf{Description} & \textbf{Example Question} & \textbf{Grounding Pool} & \textbf{Core Constraints} & \textbf{Deterministic Solver} \\
\midrule
Object Counting & Count the number of instances of an object category in the scene & ``How many windows are there in this room?'' & $\mathcal{C}_{\mathrm{scene}}$ & Target category must be countable and scene-wide & Count instances of the label \\
\addlinespace
Object Size & Estimate the longest geometric dimension of a unique target & ``What is the longest edge of this television in centimeters?'' & $\mathcal{U}_{\mathrm{scene}}$ & Target must be unique & Read the longest bounding box edge \\
\addlinespace
Absolute Distance & Measure the metric distance between the nearest points of two objects & ``What is the straight-line distance between the fireplace and the TV at their nearest points, in meters?'' & $\mathcal{U}_{\mathrm{scene}}$ & Both targets unique and distinct & Compute nearest-point distance \\
\addlinespace
Relative Distance & Identify the candidate closest to the anchor among a candidate set & ``Among the chair, coat rack, and table, which is closest to the bed at nearest points?'' & $\mathcal{U}_{\mathrm{scene}}$ & Anchor unique; candidate set valid & Return nearest object label \\
\addlinespace
Relative Direction & Standing at one object and facing another, determine the direction of a third & ``If I stand near the toilet and face the sink, in which direction is the bathtub relative to me?'' & $\mathcal{U}_{\mathrm{scene}}$ & All three roles unique and mutually distinct & Output quadrant direction label \\
\addlinespace
Room Size Estimation & Estimate the overall floor area of the room & ``Approximately how many square meters is this room?'' & N/A & Scene-level area metadata required & Read room area from metadata \\
\bottomrule
\end{tabular}
\end{adjustbox}
\end{table}

\begin{table}[h]
\centering
\caption{Validation rules and detailed descriptions for single-image tasks.}
\label{tab:single_image_tasks}
\small
\begin{adjustbox}{max width=\textwidth}
\begin{tabular}{L{2.8cm} L{3.0cm} L{4.2cm} L{2.0cm} L{3.0cm} L{2.6cm}}
\toprule
\textbf{Task} & \textbf{Description} & \textbf{Example Question} & \textbf{Grounding Pool} & \textbf{Core Constraints} & \textbf{Deterministic Solver} \\
\midrule
Single-View Relative Direction & Determine the direction of a target relative to a reference object within a single frame & ``In this image, in which direction is the sink relative to the washing machine?'' & $\mathcal{U}_{f}(v_{\min})$ & Both targets uniquely visible and distinct & Output relative direction \\
\addlinespace
Camera-Object Distance & Estimate the distance from the camera to the target object & ``In this image, what is the distance from the camera to the bed in meters?'' & $\mathcal{U}_{f}(v_{\min})$ & Target uniquely visible & Compute camera-to-object distance \\
\addlinespace
Depth Ordering & Compare which of two objects is closer to the camera & ``Which is closer to the camera, the chair or the trash can?'' & $\mathcal{U}_{f}(v_{\min})$ & Both targets uniquely visible with valid depth & Return nearer object or \textit{same} \\
\bottomrule
\end{tabular}
\end{adjustbox}
\end{table}

\begin{table}[h]
\centering
\caption{Validation rules and detailed descriptions for image-pair tasks.}
\label{tab:image_pair_tasks}
\small
\begin{adjustbox}{max width=\textwidth}
\begin{tabular}{L{2.8cm} L{3.0cm} L{4.0cm} L{2.2cm} L{2.8cm} L{2.6cm}}
\toprule
\textbf{Task} & \textbf{Description} & \textbf{Example Question} & \textbf{Grounding Pool} & \textbf{Core Constraints} & \textbf{Deterministic Solver} \\
\midrule
Inter-Camera Relative Position & Determine the relative position of Image~2's camera in Image~1's camera coordinate frame & ``When you took Image~1, where is Image~2's camera relative to you?'' & N/A & Both frame poses parseable & Output camera relative direction \\
\addlinespace
Inter-Camera Elevation & Determine the vertical relationship between two camera viewpoints & ``Compared to Image~2's camera, is Image~1's camera higher or lower?'' & N/A & Both frame poses parseable & Output \textit{higher}/\textit{lower} \\
\addlinespace
Visibility Comparison & Compare target visibility across two images & ``For the cabinet, which image shows it more clearly, Image~1 or Image~2?'' & Pair-visible object pool & Target grounded without ambiguity & Return \textit{image1}/\textit{image2}/\textit{same}/\textit{neither} \\
\addlinespace
Camera-Object Position & Determine the direction of a target object relative to the reference camera & ``When I took Image~1, in which direction is the table relative to me?'' & Ref-frame unique object pool & Target uniquely grounded in reference image & Output camera-centric direction \\
\addlinespace
Camera-Region Position & Determine the direction of a region anchor relative to the reference camera & ``When you took Image~1, in which direction is the dining area relative to you?'' & Ref-frame region anchor pool & Region phrase resolvable to an anchor & Output camera-centric direction \\
\addlinespace
Camera Motion Estimation & Determine the dominant motion direction of the camera between two frames & ``Based on this image sequence, in which direction is the camera turning?'' & N/A & Both frame poses and trajectory parseable & Output dominant motion direction \\
\addlinespace
Attribute Measurement & Compare the longest dimension of two objects & ``Which has the larger longest edge, the laptop or the oven?'' & Pair non-ambiguous object pool & Both targets distinct and uniquely bindable & Return the label of the larger object \\
\bottomrule
\end{tabular}
\end{adjustbox}
\end{table}

\subsubsection{Remarks on Degeneracy Filtering}
\label{app:B.1.3}

It is important to distinguish between the DGE's \textit{question validity judgment} and the \textit{numerical tolerance} applied during the Questioner reward stage. The former determines whether a question is deterministically solvable; the latter determines whether the model's response is sufficiently close to the DGE-computed ground truth. Accordingly, for continuous-valued tasks such as absolute distance, object size, and room size estimation, the DGE always outputs an exact geometric ground truth, while the relative error thresholds used at evaluation are defined separately in Appendix~\ref{app:C.2.2}.

Several degenerate question types are additionally suppressed by design. In object counting, categories with zero instances are excluded from the candidate pool; categories with exactly one instance yield questions of diminished discriminative value and are assigned a reduced validity weight rather than being fully rejected. For inter-camera elevation, degenerate same-level frame pairs are not actively sampled during training to avoid allocating budget to examples with negligible vertical discrimination. For depth ordering and visibility comparison, \textit{same} outcomes are retained but down-weighted in the Questioner reward, preventing the curriculum from over-concentrating on geometrically trivial edge cases.

\subsection{Automated Verification Pipeline and Extraction Prompts}
\label{app:B.2}

\subsubsection{End-to-End Verification Pipeline}
\label{app:B.2.1}

Given input $(Q, t, x)$, the automated verification pipeline executes the following sequential stages.

\textbf{Stage 1: Task normalization and input contract checking.} The noisy task name is parsed to a canonical task type, and the input is verified to satisfy the modality contract---scene-level, single-image, or image-pair---required by the current task.

\textbf{Stage 2: Context injection.} Scene metadata, frame metadata, camera poses, and visible object statistics are automatically populated from the scene identifier, image paths, and frame identifiers associated with the current sample.

\textbf{Stage 3: Structured extraction.} For tasks requiring language parsing, a lightweight language model extracts objects, candidate sets, region phrases, or image indices according to a task-specific schema. The extracted output is a structured parameter vector indexed to the task's expected fields. If certain fields remain null after extraction, a heuristic fallback attempts to resolve them from the grounded candidate set before proceeding.

\textbf{Stage 4: Canonicalization and sanitization.} The extracted parameter vector undergoes lowercasing, label normalization against the scene's grounded object pool, candidate pool constraint enforcement, and heuristic completion. Region phrases additionally pass through an ontology-based anchor resolution step.

\textbf{Stage 5: Rule-based rejection.} If any extracted field is empty, any label falls outside the permitted grounded pool, roles conflict, or candidate lists are structurally invalid, the DGE immediately returns a rejection decision together with a task-specific invalidity reason, without invoking any downstream geometric operator.

\textbf{Stage 6: Deterministic rubric execution.} For questions passing all prior checks, the DGE invokes the geometry operators prescribed by the task's rubric in sequence and synthesizes the final ground-truth answer. The pipeline additionally records intermediate geometric states---including key point coordinates, relative pose matrices, and depth projections---to support downstream qualitative analysis and error attribution.

\subsubsection{Prompted Entity Extraction}
\label{app:B.2.2}

Rather than relying on a single free-form parser, the DGE specifies an explicit extraction schema for each task and instructs the extraction model to output only the structured fields. Scene-level and single-image tasks share the following common prompt prefix. And image-pair tasks use a more camera-relation-oriented prefix.

\begin{promptbox}[Entity Extraction Prompt -- Scene and Single-Image Tasks]{black}
Role: Expert Entity Extractor for 3D Scenes.\\
Instruction: Extract objects from the question exactly as they appear.\\
Rules:\\
- Use lowercase and singular forms.\\
- Keep original spelling; do not paraphrase or replace with synonyms.\\
- If missing or uncertain, output null.\\
- Output only the extraction result in the required format.
\end{promptbox}

\begin{promptbox}[Entity Extraction Prompt -- Image-Pair Tasks]{black}
Role: Expert Entity Extractor for spatial reasoning.\\
Instruction: Extract entities exactly as asked in the question.\\
Rules:\\
- Output lowercase.\\
- Keep nouns only.\\
- If uncertain, output null.\\
- Output only the required format.
\end{promptbox}

Each task appends its own schema specification on top of this prefix. Some representative templates are provided below.

\begin{promptbox}[Extraction Schema -- Object Counting]{black}
Task: Extract the object category being counted.\\
Important:\\
- Extract only the queried category.\\
- Use lowercase.\\
- If missing or uncertain, output \texttt{null}.\\
Format: category\_name\\
Question: \{question\}\\
Extraction:
\end{promptbox}

\begin{promptbox}[Extraction Schema -- Single-View Relative Direction]{black}
Task: Extract two objects for relative direction in one image.\\
Important:\\
- The answer means the target object relative to the reference object.\\
- Output the reference object first and the target object second.\\
- If either object is missing or uncertain, output \texttt{null} for that slot.\\
Format: reference\_object, target\_object\\
Question: \{question\}\\
Extraction:
\end{promptbox}

For image-pair spatial questions, we separate camera reference parsing from semantic entity extraction. The observer image index is first inferred from the question (e.g., “when taking Image~1”) and stored as a camera-reference variable. Conditioned on this, the extraction module only identifies the queried object or region, reducing ambiguity and yielding simpler, more robust prompts across diverse camera-centered questions.

\begin{promptbox}[Extraction Schema -- Camera-Object Position]{black}
Task: Extract the target object phrase relative to the camera.\\
Important:\\
- The observer image index (Image~1 or Image~2) is resolved separately and should NOT be extracted here.\\
- Extract only the queried object phrase.\\
- Keep the object phrase concise and noun-based.\\
- If the target object cannot be determined from the question, output \texttt{null}.\\
Format: target\_object\\
Question: \{question\}\\
Extraction:
\end{promptbox}

\begin{promptbox}[Extraction Schema -- Camera-Region Position]{black}
Task: Extract the target region phrase relative to the camera.\\
Important:\\
- The observer image index (Image~1 or Image~2) is resolved separately and should NOT be extracted here.\\
- If the question refers to an abstract region such as a sleeping area, bathroom area, kitchen area, or living area, keep that phrase
verbatim.\\
- Do NOT replace the region phrase with an object label.\\
- Extract only the queried region phrase.\\
- If the target region cannot be determined from the question, output \texttt{null}.\\
Format: region\_name\\
Question: \{question\}\\
Extraction:
\end{promptbox}

Not all tasks require LLM-based extraction. Room size estimation, inter-camera relative position, inter-camera elevation, and camera motion estimation are fully determined by scene or camera pose parameters and therefore skip the extraction stage entirely.

\subsubsection{Questioner Prompt Templates Used in Round~1}
\label{app:B.2.3}

The Questioner in Round~1 employs three task-conditioned prompt templates corresponding to scene-level, single-image, and image-pair input modes. Curly-brace placeholders denote runtime-populated content. The three templates share the same structural organization but differ in their grounding scope: the scene-level template mandates whole-scene evidence; the single-image template restricts all grounding to the current frame's visible content; and the image-pair template emphasizes cross-frame comparison and requires explicit Image~1 or Image~2 references when the task focuses on a specific viewpoint.

\begin{promptbox}[Questioner Prompt -- Scene-Level Template]{black}
\# SYSTEM ROLE:\\
You are an expert 3D scene analyst and question designer.\\
Assigned task type: \{task\_display\_name\}.\\
\\
\# TASK GOAL:\\
\{task\_goal\}\\
\\
\# AVAILABLE CONTEXT:\\
\{task\_context\_sections\}\\
\\
\# TASK-SPECIFIC VALIDITY GUIDE:\\
\{task\_candidate\_guidance\}\\
\\
\# YOUR JOB:\\
1. Write one grounded whole-scene observation, then exactly one scene-level question for the assigned task.\\
2. Make the observation flow from global scene layout to the local target so it naturally supports the question.\\
3. Keep the observation detailed and spatially grounded rather than short or list-like.\\
4. Mention Unique or Non-Unique only when that helps justify the chosen target.\\
\\
\# HARD CONSTRAINTS:\\
- Use whole-scene evidence, not a single sampled frame.\\
- Copy object labels exactly from the provided label list.\\
- For tasks requiring unique objects, use labels marked (Unique).\\
- Avoid generic, list-only, or weakly grounded observations.\\
\\
\# OUTPUT FORMAT:\\
\textless observation\textgreater...\textless/observation\textgreater\\
\textless question\textgreater...\textless/question\textgreater
\end{promptbox}

\begin{promptbox}[Questioner Prompt -- Single-Image Template]{black}
\# SYSTEM ROLE:\\
You are an expert single-image spatial reasoning question designer.\\
Assigned task type: \{task\_display\_name\}.\\
\\
\# TASK GOAL:\\
\{task\_goal\}\\
\\
\# AVAILABLE CONTEXT:\\
\{task\_context\_sections\}\\
\\
\# TASK-SPECIFIC VALIDITY GUIDE:\\
\{task\_candidate\_guidance\}\\
\\
\# YOUR JOB:\\
1. Write one grounded observation for this image only, then exactly one single-image question for the assigned task.\\
2. Make the observation flow from overall image layout to the local target.\\
3. Use only relations visible in this image; keep the observation detailed rather than list-like.\\
4. Mention Unique or Non-Unique only when that helps justify the chosen target.\\
\\
\# HARD CONSTRAINTS:\\
- Use only evidence from the current image.\\
- Copy object labels exactly from the provided label list.\\
- Prefer (Unique) labels when ambiguity would otherwise be high.\\
- Avoid generic, list-only, or cross-image reasoning.\\
\\
\# OUTPUT FORMAT:\\
\textless observation\textgreater...\textless/observation\textgreater\\
\textless question\textgreater...\textless/question\textgreater
\end{promptbox}

\begin{promptbox}[Questioner Prompt -- Image-Pair Template]{black}
\# SYSTEM ROLE:\\
You are an expert two-image spatial reasoning question designer.\\
Assigned task type: \{task\_display\_name\}.\\
\\
\# TASK GOAL:\\
\{task\_goal\}\\
\\
\# AVAILABLE CONTEXT:\\
\{task\_context\_sections\}\\
\\
\# TASK-SPECIFIC VALIDITY GUIDE:\\
\{task\_candidate\_guidance\}\\
\\
\# YOUR JOB:\\
1. Write one grounded comparative observation for the image pair, then exactly one image-pair question for the assigned task.\\
2. Make the observation flow from pair-level relation to the local target.\\
3. If the task focuses on one image, explicitly say Image~1 or Image~2.\\
4. Mention Unique, Non-Unique, or shared visibility only when that helps justify the chosen target.\\
\\
\# HARD CONSTRAINTS:\\
- Use only evidence grounded in the provided image pair.\\
- Copy object labels exactly from the provided label list.\\
- Explicitly state Image~1 or Image~2 when the task requires a concrete image reference.\\
- Avoid generic, weakly grounded, or repetitive observations.\\
\\
\# OUTPUT FORMAT:\\
\textless observation\textgreater...\textless/observation\textgreater\\
\textless question\textgreater...\textless/question\textgreater
\end{promptbox}

\subsubsection{Candidate-Pool Sanitization and Heuristic Fallback}
\label{app:B.2.4}

Before passing extracted results to the solver, two normalization layers are applied. The first is \textbf{lexical sanitization}, which includes case normalization, compact matching for compound labels (e.g., mapping \textit{nightstand} to \textit{night stand}), phrase-level matching, and word-level mapping after stopword removal. The second is \textbf{pool-aware sanitization}: if an extracted label does not appear in the grounded pool for the current sample, the system first attempts to remap it to the nearest canonical label; if this fails, the field is set to null within the heuristic candidate set, triggering downstream rejection.

For camera-region position tasks, the DGE maintains a \textbf{region-object ontology}. When a question supplies an abstract region phrase such as \textit{sleeping area}, the system does not immediately reject the question; it instead retains the phrase and resolves it to a spatially grounded anchor object via the ontology, the available anchor candidate list, and, when necessary, an auxiliary language model call. The question is rejected only if this resolution ultimately fails.

\subsubsection{Unified Response Object}
\label{app:B.2.5}

Once a question passes the full pipeline, the DGE returns a unified structured response object containing at minimum: a validity flag, a deterministic supervision signal, structured parsing results, and a failure summary including the error code, failure stage, and validation result. For invalid questions, the system additionally constructs a diagnostic summary used by the explanation judge in Round~2. This unified interface enables the DGE to simultaneously fulfill two responsibilities: providing noise-free geometric ground truth for valid questions, and providing task-semantically interpretable rejection evidence for invalid ones.

\section{Algorithmic Details of Spatial-Grounded Policy Co-Evolution}
\label{app:C}

This appendix provides a detailed algorithmic exposition of the Spatial-Grounded Policy Co-Evolution mechanism introduced in Section~\ref{Spatial-Grounded Policy Co-Evolution}, with the aim of ensuring full reproducibility of the co-evolution training procedure. We proceed as follows: the task-adaptive scheduling formulation (Appendix~\ref{app:C.1}), and detailed reward specifications for both the Questioner and the Solver (Appendices~\ref{app:C.2} and~\ref{app:C.3}).

\subsection{Task-Adaptive Scheduling}
\label{app:C.1}

\subsubsection{Feasible-Task Inference}
\label{app:C.1.1}

The feasible task set $\mathcal{T}^{\mathrm{feasible}}_s$ is not manually specified but is inferred online from the DGE environment summary $\Sigma_s$ of the current sample:
\begin{equation}
\mathcal{T}^{\mathrm{feasible}}_s
=
\{k\in\mathcal{T}\mid \phi_k(\Sigma_s)=1\},
\end{equation}
where $\phi_k$ is the feasibility predicate for task $k$. For \textbf{scene-level tasks}: object counting is feasible if at least one non-unique label exists; absolute distance requires at least two unique labels; object size requires at least one; room size estimation is always feasible; relative distance requires at least four unique labels; and relative direction requires at least three. For \textbf{single-image tasks}: single-view relative direction and depth ordering each require at least two uniquely visible objects; camera-object distance requires at least one. For \textbf{image-pair tasks}: inter-camera relative position and camera motion estimation are feasible whenever both frame poses are available; inter-camera elevation is feasible only when the height difference is non-degenerate; visibility comparison is feasible if the pair visibility pool is non-empty; camera-object position requires at least one uniquely visible label in the reference frame; camera-region position requires at least one resolvable region anchor; and attribute measurement requires at least two labels in the pair's non-ambiguous object pool.

\subsubsection{Scheduler-Adjusted Historical Accuracy}
\label{app:C.1.2}

The scheduler maintains three statistics for each task $k$: historical sample count $N_k$, cumulative raw accuracy $S_k$, and a calibrated accuracy sum $S_k^{\mathrm{sched}}$ used specifically for scheduling~\cite{chen2025self}. The calibrated accuracy applies task-specific difficulty normalization:
\begin{equation}
a_k^{\mathrm{sched}}
=
\mathrm{clip}\!\left(\frac{a_k}{\tau_k},\, 0,\, 1\right),
\qquad
\tau_k
=
\begin{cases}
0.50, & k \in \mathcal{T}_{\mathrm{num}},\\
1.00, & \text{otherwise,}
\end{cases}
\end{equation}

where $\mathcal{T}_{\mathrm{num}}$ denotes the set of all numeric-output tasks. Setting $\tau_k=0.5$ for numeric tasks reflects the fact that their accuracy is a smooth continuous metric rather than a hard exact-match score~\cite{chen2024spatialvlm}; a calibrated value exceeding $0.5$ thus serves as a natural indicator of adequate mastery.

\subsubsection{Task Sampling from Smoothed Accuracy}
\label{app:C.1.3}

The smoothed accuracy $\bar{a}_k$ is computed with a pseudo-count prior to mitigate estimation instability during early training:
\begin{equation}
\bar{a}_k
=
\frac{S_k^{\mathrm{sched}} + a_0 n_0}{N_k + n_0},
\qquad
a_0 = 0.35,\quad n_0 = 2.0.
\end{equation}
Task sampling weights are then computed as $w_k = \max(\delta,\, 1-\bar{a}_k)$ with $\delta = 0.05$, assigning higher weight to tasks with lower historical accuracy while ensuring well-mastered tasks are not entirely excluded. The final sampling probability within the feasible task set is
\begin{equation}
p_k
=
\frac{w_k}{\displaystyle\sum_{j\in\mathcal{T}^{\mathrm{feasible}}_s} w_j},
\qquad
k\in\mathcal{T}^{\mathrm{feasible}}_s.
\end{equation}

\subsubsection{Deduplication-Aware Statistics}
\label{app:C.1.4}

As noted in Section~\ref{3.3.5:GRPO} Round~1 samples multiple candidate questions per visual context. Before entering Round~2, an intra-context deduplication step is applied; deduplication operates strictly within each source context and does not merge questions across distinct scene samples.

Each candidate question is assigned a \textbf{semantic signature} determined jointly by its canonical task type and the structured parameters extracted by the DGE, rather than by the raw question string. Fields such as object pairs, candidate sets, and image references are normalized before signature construction, including lowercasing, canonical image index mapping, and sorting of unordered object pairs. For tasks with explicit image roles such as inter-camera relative position and camera-object position, reference and target image roles are encoded in the signature. Exactly one representative question per unique semantic signature is retained for Round~2. To preserve curriculum consistency despite deduplication, each retained representative carries a \textbf{duplicate weight} equal to its signature's frequency in the original rollout; the scheduler statistics $(N_k, S_k, S_k^{\mathrm{sched}})$ are updated using this weight rather than the deduplicated question count. If a retained question is later judged invalid before answer evaluation, its scheduler-facing accuracy is set to zero, while its duplicate weight is still counted in the effective statistics.

\subsection{Questioner-Side Reward}
\label{app:C.2}

\subsubsection{Final Reward Form}
\label{app:C.2.1}

The Questioner reward in the full implementation takes the form
\begin{equation}
r_Q
=
\begin{cases}
-1, & \text{if a severe structural failure occurs},\\
0.1\, f_{\mathrm{fmt}} + 0.9\, f_{\mathrm{valid}}\, f_{\mathrm{obs}}, & \text{otherwise,}
\end{cases}
\end{equation}
where $f_{\mathrm{fmt}}$ measures adherence to the expected observation-question template, $f_{\mathrm{valid}}$ reflects the DGE's geometric validity and non-degeneracy assessment, and $f_{\mathrm{obs}}$ is assigned by a lightweight text judge assessing the quality of reasoning in the observation description.

\subsubsection{Validity Factor}
\label{app:C.2.2}

$f_{\mathrm{valid}}$ functions first as a binary gate: the DGE rejection yields $f_{\mathrm{valid}}=0$; a standard valid question yields $f_{\mathrm{valid}}=1$. Task-specific downward adjustments are applied to questions that are valid yet carry limited discriminative value. In object counting, a true count of zero renders the question invalid; a count of exactly one is assigned $f_{\mathrm{valid}}=0.5$. In depth ordering, a \textit{same} outcome yields $f_{\mathrm{valid}}=0.5$. In inter-camera elevation, a same-level outcome is assigned $f_{\mathrm{valid}}=0$. In visibility comparison, outcomes of \textit{same} or \textit{neither} are assigned $f_{\mathrm{valid}}=0.5$. This design suppresses the Questioner from repeatedly sampling edge-case questions that, while formally valid, offer limited geometric learning signal.

\subsubsection{Observation-Based Judgment}
\label{app:C.2.3}

Observation quality is scored by a GPT-OSS-120B~\cite{agarwal2025gpt} text judge that assesses whether the observation constitutes a high-quality visual-to-question transition rather than re-evaluating question correctness. The verbatim prompt is provided below.

\begin{promptbox}[Observation Quality Judge Prompt]{black}
You are evaluating the quality of a task-conditioned observation written by a 3D spatial reasoning model.\\
\\
\#\#\# OBSERVATION: \{observation\}\\
\#\#\# QUESTION TYPE CHOSEN: \{q\_type\_str\}\\
\#\#\# QUESTION GENERATED: \{raw\_question\}\\
\\
\#\#\# CONTEXT:\\
The ideal Observation has a global-to-local flow: (1) summarize the overall scene or image-pair layout; (2) narrow to the local objects, region, or relation most relevant to the task; (3) lead naturally into the exact target of the question; (4) explain Unique/Non-Unique status when useful. The Observation need not describe every visible object.\\
You are NOT grading question correctness---you are grading whether the Observation is a good visual-spatial lead-in for the question. Judge holistically. Do NOT apply fixed word-count or clause-count rules.\\
\\
\#\#\# YOUR TASK:\\
Evaluate on: (1) Groundedness, (2) Global caption quality, (3) Local focus quality, (4) Transition quality, (5) Support quality, (6) Sample-specificity.\\
\\
\#\#\# SCORING RULES:\\
1.0 -- Clearly grounded, meaningful overall caption, narrows to the task-relevant target, makes the question feel natural, sample-specific.\\
0.6 -- Grounded and useful, but one major component is weaker.\\
0.3 -- Shows real visual effort but is fragmentary, generic, or only loosely connected to the question.\\
0.0 -- Empty, template-like, contradictory, hallucinated, or lacks visual-spatial content.\\
\\
Think step by step. Output ONLY:\\
\textless score\textgreater[1.0 / 0.6 / 0.3 / 0.0]\textless/score\textgreater
\end{promptbox}

The judge is invoked only when the question is valid, the format is correct, and the observation is non-empty without inner-tag contamination. To prevent the coupled reward from collapsing due to an excessively conservative judge score, a minimal lower bound is imposed: $f_{\mathrm{obs}} = \max(0.1,\, f_{\mathrm{obs}}^{\mathrm{judge}})$. For cases with no observation, an invalid question, or a format violation, $f_{\mathrm{obs}}$ is set to $0$ directly.

\subsubsection{Representative High- and Low-Score Observations}
\label{app:C.2.4}

We provide two illustrative examples consistent with the scoring criteria, using abstract object names to avoid overlap with main-paper case studies.

\begin{promptbox}[High-Score Observation Example (score 1.0 or 0.6)]{black}
Across the two images, the overall room layout remains stable while the camera shifts slightly toward the right side of the scene. Large furniture still anchors the perimeter, and the open floor in the middle makes the camera change easy to track. Narrowing to Image~2, the chair is isolated near the left side of the current viewpoint with clear free space between the camera and the chair, making it the most natural grounded target for a camera-relative direction question.
\end{promptbox}

\begin{promptbox}[Low-Score Observation Example (score 0.3 or 0.0)]{black}
There is a chair in the room. The question is about the chair.
\end{promptbox}

The high-score example first describes the pair-level global structure, then narrows to the task-relevant local target, and transitions naturally to a camera-relative direction question. The low-score example provides neither a global description nor a justification for target selection, and is essentially a restatement of the target label.

\subsection{Solver-Side Reward}
\label{app:C.3}

\subsubsection{Round~2 Format and Reward Decomposition}
\label{app:C.3.1}

For valid questions, the Solver receives the following prompt:

\begin{promptbox}[Solver Prompt -- Valid Question]{black}
Carefully examine the images and answer the following question step by step.\\
Write your reasoning in plain text, keeping each step grounded in the visual evidence.\\
Avoid redundant or repetitive descriptions.\\
Only the final line should contain the answer in \textless answer\textgreater...\textless/answer\textgreater\ format.\\
\\
Question: \{question\}\\
\\
Response:
\end{promptbox}

For invalid questions, the Solver generates a structured explanation of the invalidity reason:

\begin{promptbox}[Solver Prompt -- Invalid Question]{black}
You are an expert spatial reasoning analyst.\\
The following question has already been judged invalid for the assigned task. Your job is to explain why it is invalid based on the task validity rules below.\\
\\
Assigned task type: \{task\_display\_name\}\\
Question: \{question\}\\
\{task\_reference\_text\}\\
\\
Explain step by step why this question is invalid. Ground each step in the task validity rules and the simulator-backed diagnostic context.\\
Focus on the main issue, such as: using object labels outside the grounded pool; using an ambiguous or non-unique object; using the wrong number of candidates or roles; reusing the same object in conflicting roles; or asking in a way that does not match the assigned task schema.\\
\\
Provide the final conclusion inside \textless answer\textgreater\ tags.\\
Example: \textless answer\textgreater The target object is not uniquely grounded.\textless/answer\textgreater\\
\\
Response:
\end{promptbox}

The Solver reward in Round~2 is
\begin{equation}
r_A
=
\begin{cases}
-1, & \text{if } Q \text{ is valid and the response has a hard format failure},\\
0.1\, f_{\mathrm{fmt}} + 0.9\, f_{\mathrm{acc}}, & \text{if } Q \text{ is valid},\\
0.1\, f_{\mathrm{fmt}} + 0.9\, f_{\mathrm{explain}}, & \text{if } Q \text{ is invalid,}
\end{cases}
\end{equation}
where $f_{\mathrm{fmt}}\in\{-1,0,1\}$ equals $1$ if exactly one \texttt{<answer>} tag is present with no other XML/HTML-style tags, $0$ if the tag is absent, and $-1$ if multiple tags or other XML/HTML-style tags appear. A hard format failure on a valid question triggers an overall penalty of $-1$; for invalid-question explanations, the format score enters the reward linearly without this hard penalty.

\subsubsection{Accuracy Scoring for All 16 Tasks}
\label{app:C.3.2}

Rather than approximating ground truth via majority-vote consensus, the implementation scores each sample directly against the deterministic ground truth produced by the DGE. For continuous-valued tasks, accuracy is computed via a relative-error threshold grid~\cite{li2025spatialladder}:
\begin{equation}
d_{\mathrm{rel}}(\hat{y}, y)
=
\frac{|\hat{y}-y|}{\max(y,\varepsilon)},
\quad
A_{\mathrm{rel}}(\hat{y},y)
=
\frac{1}{|\mathcal{C}_{\mathrm{rel}}|}
\sum_{c\in\mathcal{C}_{\mathrm{rel}}}
\mathbb{I}\!\left[
d_{\mathrm{rel}}(\hat{y},y)\le 1-c
\right],
\quad
\mathcal{C}_{\mathrm{rel}}
=
\operatorname{LinSpace}(0.50,\,0.95,\,11),
\end{equation}
with $\varepsilon = 10^{-9}$. This constructs 11 equally-spaced confidence points mapped into tolerance thresholds ranging from coarse to strict. Object size, absolute distance, room size estimation, and camera-object distance all use $A_{\mathrm{rel}}$ as $f_{\mathrm{acc}}$.

Object counting uses a piecewise absolute-error reward: exact match scores $1.0$; error $\le 1$ scores $0.3$; error $\le 2$ scores $0.1$; all other cases score $0$. Relative distance is evaluated by exact label match. Relative direction first normalizes direction aliases and decomposes compound directions into set form, then applies set consistency comparison. Single-view relative direction uses direction-set matching with partial-credit rules for subset matches. Depth ordering is a three-class task (object 1, object 2, or same) evaluated by exact match after label normalization. Inter-camera relative position, camera-object position, and camera-region position apply camera-centric direction-set matching; camera motion estimation uses the same rule, mapping clockwise/counterclockwise to right/left in the normalization step. Inter-camera elevation, visibility comparison, and attribute measurement are all evaluated by exact match after label normalization.

\subsubsection{Invalid-Question Explanation Scoring}
\label{app:C.3.3}

When the DGE rejects a question, the system converts the structured rejection evidence into an explanation-based training signal rather than discarding the sample. Explanation quality is scored with $f_{\mathrm{explain}} \in \{1.0,\, 0.6,\, 0.3,\, 0.0\}$, where a score of $1.0$ requires correctly identifying the DGE's primary invalidity reason and grounding it in task rules, the object pool, extraction failure, non-unique reference, or a validation constraint; $0.6$ indicates the main failure is mostly captured but an important detail is missing; $0.3$ applies when only part of the problem is identified; and $0.0$ penalizes contradictions or complete failure to explain the invalidity.

The judge receives the model's explanation alongside authoritative DGE diagnostic signals including the final invalid reason, error code, failure stage, extraction field resolution status, and validation issues, assessing whether the explanation faithfully reflects the DGE's structured conclusions. The verbatim prompt is as follows.

\begin{promptbox}[Invalid-Question Explanation Judge Prompt]{black}
You are judging whether a model correctly explained why a spatial reasoning question is invalid.\\
The simulator diagnostics below are authoritative deterministic ground truth from the real validation pipeline.\\
Your job: check whether the model explanation captures the simulator's actual failure reason.\\
Prefer the simulator's final conclusion over fluent but unsupported explanations.\\
\\
\#\#\# ASSIGNED TASK TYPE: \{task\_type\}\\
\#\#\# ORIGINAL QUESTION: \{question\}\\
\#\#\# TASK REFERENCE: \{task\_reference\}\\
\#\#\# SIMULATOR INVALID SIGNALS: \{simulator\_reference\}\\
\#\#\# MODEL EXPLANATION: \{explanation\}\\
\\
\#\#\# SCORING RUBRIC:\\
1.0 -- Correctly identifies the simulator's main invalid reason and grounds it in task rules, grounded object pool, extraction failure, non-unique reference, or validation issue.\\
0.6 -- Mostly correct on the main failure, but misses an important supporting detail or states it too vaguely.\\
0.3 -- Catches only part of the problem, or gives a generic reason that weakly overlaps with the diagnostics.\\
0.0 -- Wrong, contradicts the simulator diagnostics, or fails to explain the invalidity.\\
\\
Return ONLY one score: 0 / 0.3 / 0.6 / 1.0
\end{promptbox}

The key property of this mechanism is that the training signal for invalid questions is anchored to the DGE's structured rejection evidence rather than to inter-model consensus. Even geometrically unsolvable samples therefore provide valuable supervision: they teach the model which questions should not be asked, why, and how to correctly attribute the failure to the appropriate geometric or task constraint.

\section{Implementation and Experimental Details}
\label{app:D}

This appendix supplements Section~\ref{4:Experimental} with complete implementation details covering the two-stage training configuration, the composition of SFT and RL data, and the experimental scope of the ablation and analysis studies.

\subsection{Training Configuration}
\label{app:D.1}

\subsubsection{Backbone Model and Training Configuration}
\label{app:D.1.1}
Unless otherwise specified, the main experiments evaluate both \textbf{Qwen2.5-VL-7B-Instruct} and \textbf{Qwen2.5-VL-3B-Instruct} as backbone models, while all other experiments default to \textbf{Qwen2.5-VL-7B-Instruct}.

The default training paradigm of SpatialEvo is online GRPO reinforcement learning, applied directly without an SFT warm-start. The paradigm comparison in Table~\ref{tab:paradigm} additionally includes an SFT baseline for reference; its hyperparameters are reported in Table~\ref{tab:hparam_sft} and remain unchanged from the configuration described therein. The complete hyperparameter configurations for both stages are reported in Tables~\ref{tab:hparam_sft} and~\ref{tab:hparam_rl}.

\begin{table}[h]
\centering
\begin{minipage}[t]{0.46\textwidth}
\centering
\caption{Hyperparameters for the offline SFT warm-start stage.}
\label{tab:hparam_sft}
\small
\begin{tabular}{ll}
\toprule
\textbf{Hyperparameter} & \textbf{Value} \\
\midrule
global\_batch\_size & 96 \\
gradient\_accumulation\_steps & 6 \\
micro\_batch\_size & 1 \\
bf16 & true \\
data\_seed & 42 \\
gradient\_checkpointing & true \\
attn\_implementation & flash\_attn \\
lr\_scheduler\_type & cosine \\
warmup\_ratio & 0.1 \\
learning\_rate & $2\times10^{-5}$ \\
num\_train\_epochs & 1 \\
max\_pixels & 802,816 \\
min\_pixels & 3,136 \\
sequence\_parallel & true \\
tensor\_parallel\_size & 2 \\
\bottomrule
\end{tabular}
\end{minipage}
\hfill
\begin{minipage}[t]{0.50\textwidth}
\centering
\caption{Hyperparameters for the online GRPO reinforcement learning stage.}
\label{tab:hparam_rl}
\small
\begin{tabular}{ll}
\toprule
\textbf{Hyperparameter} & \textbf{Value} \\
\midrule
num\_generations                & 4 \\
per\_device\_train\_batch\_size & 1 \\
gradient\_accumulation\_steps   & 4 \\
bf16                            & true \\
data\_seed                      & 42 \\
gradient\_checkpointing         & true \\
attn\_implementation            & flash\_attn \\
num\_train\_epochs              & 4 \\
actor\_learning\_rate           & $1\times10^{-6}$ \\
KL coefficient ($\beta$)        & $10^{-2}$ \\
temperature                     & 1.0 \\
top-$p$                         & 1.0 \\
max\_pixels                     & 150{,}528 \\
min\_pixels                     & 100{,}352 \\
tensor\_parallel\_size          & 2 \\
\bottomrule
\end{tabular}
\end{minipage}
\end{table}

The RL stage is implemented using the EasyR1 framework and runs on 1 node $\times$ 8 H800 (80\,GB) GPUs with a two-round architecture: Round~1 handles Questioner-side question generation and Round~2 handles Solver-side answer generation. With $n_{\mathrm{rollout}}=4$ maintained across both rounds, each source context yields at most $4\times4=16$ candidate question-answer rollout chains in the ideal upper bound without deduplication. The SFT baseline is implemented using the MS-Swift framework and trained on 4 nodes $\times$ 8 H800 (80\,GB) GPUs with sequence parallelism and tensor model parallelism (size~$=2$) under a full-shard training configuration.

\subsubsection{Auxiliary Language Models}
\label{app:D.1.2}

All auxiliary language model calls within the framework---DGE entity and region extraction (Appendix~\ref{app:B.2}), Round~1 observation quality judging (Appendix~\ref{app:C.2}), and Round~2 invalid-question explanation judging (Appendix~\ref{app:C.3})---are unified to a single \textbf{GPT-OSS-120B}~\cite{agarwal2025gpt} text backend. All auxiliary calls are pure text-only operations and introduce no additional vision backbone. This unification controls system complexity, avoids uncertainty from additional visual modules, and ensures all judges and parsers share the same language prior, reducing system-level heterogeneity.

\subsection{Online RL Data Composition}
\label{app:D.2}

Online RL operates over a pre-filtered multi-source visual context pool rather than sampling directly from raw video streams. The pool contains scene-level multi-frame, image-pair, and single-image contexts drawn from the training splits of ScanNet, ScanNet++, and ARKitScenes. Crucially, the policy model receives raw image contexts as input; both question and answer generation are performed online during training.

Quality filtering ensures that scene-level contexts have sufficiently high grounded visible object counts and low zero-visibility ratios; image-pair contexts contain at least three shared visible objects across frames and at least five per frame; and single-image contexts include at least six visible objects. During training, for each video we sample a limited number of contexts per modality (no more than three), preventing excessive data redundancy while maintaining diverse and informative visual inputs. After this process, the final context pool contains 4,365 contexts, with the composition shown in Table~\ref{tab:rl_pool}. The online RL context pool is balanced by modality in proportion to the number of supported task types, yielding an approximate 6:7:3 ratio for scene-level, image-pair, and single-image inputs.

\begin{table}[h]
\centering
\caption{Composition of the online RL context pool by modality and data source.}
\label{tab:rl_pool}
\small
\begin{tabular}{lrrrr}
\toprule
\textbf{Input Modality} & \textbf{ScanNet} & \textbf{ScanNet++} & \textbf{ARKitScenes} & \textbf{Total} \\
\midrule
Scene-level  & 846 & 171 & 620 & 1,637 \\
Image-pair   & 884 & 298 & 728 & 1,910 \\
Single-image & 281 & 97  & 440 & 818   \\
\midrule
\textbf{Total} & \textbf{2,011} & \textbf{566} & \textbf{1,788} & \textbf{4,365} \\
\bottomrule
\end{tabular}
\end{table}
\subsection{Ablation and Analysis Experiment Details}
\label{app:D.3}

\paragraph{Majority-Vote Pseudo-Label Ablation}
\label{app:ablation_vote_pseudo_gt}

Unless otherwise specified, all ablation experiments share exactly the same training configuration as the main method; the sole variable is the source of Round~2 supervision.

In the \textit{w/o Physical Grounding} ablation, we replace DGE-derived ground truth with pseudo labels obtained by voting over the model's own sampled responses. Following Round~1 deduplication, all Round~2 responses per question are aggregated using task-appropriate strategies: majority vote for discrete and direction-based tasks, integer-rounded plurality for counting, and the median of the largest tolerance-clustered group for continuous values. For invalid-question cases, free-form explanations are mapped to canonical failure labels and the plurality is used. If no consensus is reachable, the pseudo label is left empty and the content reward is set to zero.

The Round~2 reward retains the same format--content decomposition as the main method, with the content term computed against these vote-based pseudo labels instead of geometric ground truth. This ablation thereby isolates the contribution of physically grounded supervision.

\paragraph{Reproduced baselines.}
To ensure a fair comparison with static-dataset methods in Section~\ref{5.1: Online Evolution}, we reproduce SpatialLadder under a unified training framework, fine-tuned from the same Qwen2.5-VL-3B backbone on the official 26K corpus under identical SFT settings, with all remaining hyperparameters following our configuration. For SpaceR and Spatial-SSRL, we directly evaluate their publicly released checkpoints based on Qwen2.5-VL-3B without further reproduction. For the GRPO comparison, we train on the SpatialLadder dataset with a rollout multiplicity of 8 to match the original paper, while all remaining hyperparameters follow our method as reported in Table~\ref{tab:hparam_rl}.

\paragraph{SpatialEvo controlled configuration.}
In Section~\ref{5.1: Online Evolution}, SpatialEvo is restricted to the training split of the ScanNet dataset, comprising 992 scenes, and six task categories: object counting, object size, room size estimation, absolute distance, relative distance, and relative direction. All RL hyperparameters (except for the number of training epochs) follow Table~\ref{tab:hparam_rl}, with the number of epochs determined dynamically based on data generation. Online RL training runs until approximately 20K non-redundant question-answer pairs have been generated---corresponding to roughly 12 training epochs---which can be tracked using the method described in Section~\ref{app:C.1.4}. The SFT comparison uses an offline corpus collected from the same online training trajectory, organized into multiple-choice (MCQ) or numerical (NQ) format, consistent with other datasets.


\section{Supplementary Qualitative Cases}
\label{app:E}

\begin{figure}[t]
\centering
\includegraphics[width=0.98\textwidth,
    trim={0pt 0pt 0pt 0pt},
    clip]{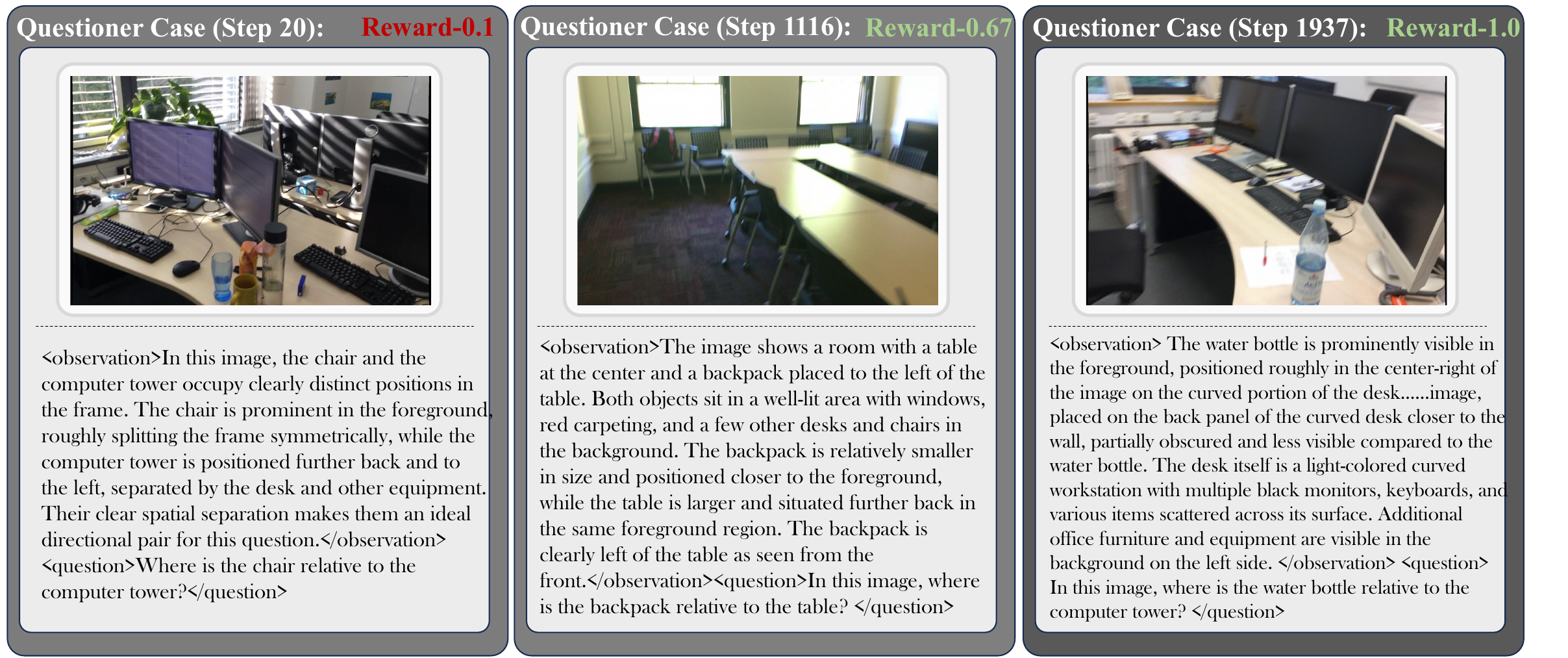}
\caption{Evolution of the Questioner across training steps on the single-view relative direction task. Observations become increasingly detailed and spatially grounded, with generated questions more tightly coupled to visual evidence.}
\label{fig:questioner_case}
\vspace{-1.2em}
\end{figure}

\begin{figure}[t]
\centering
\includegraphics[width=0.98\textwidth,
    trim={0pt 0pt 0pt 0pt},
    clip]{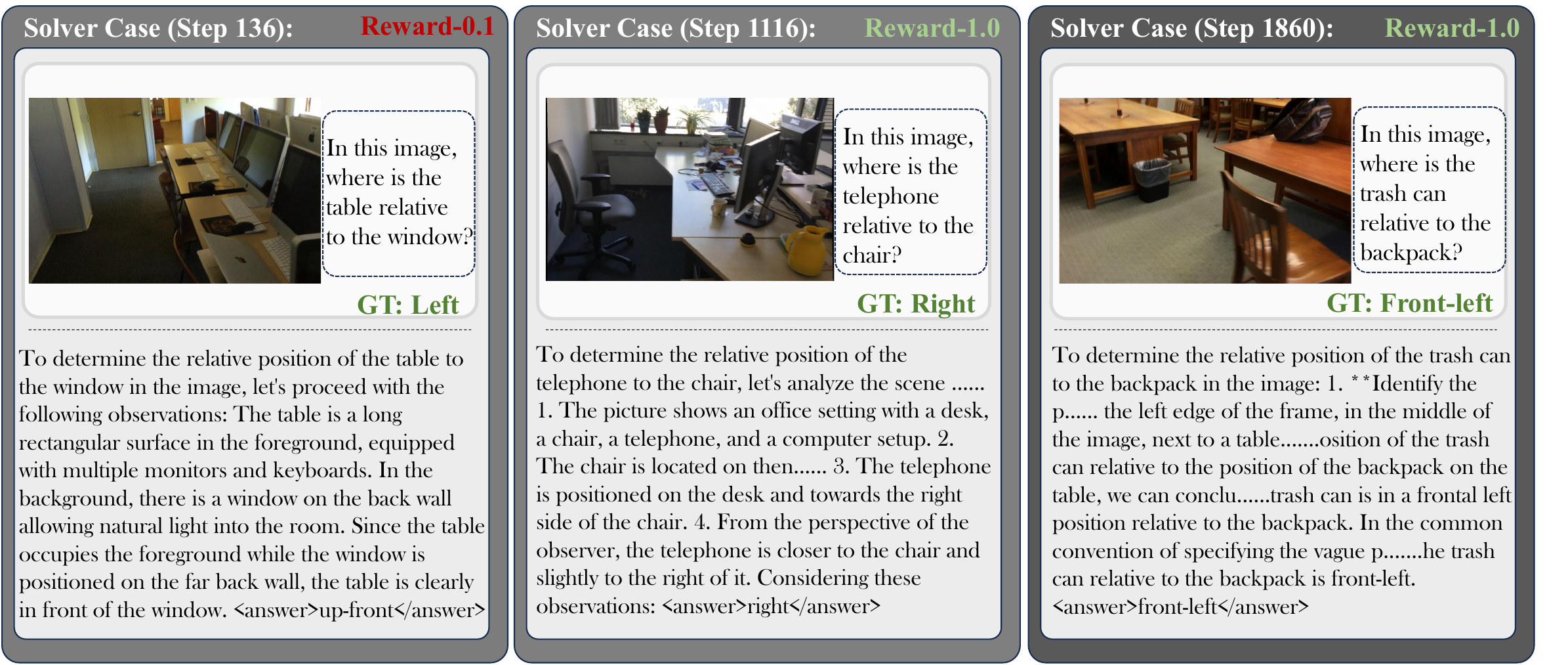}
\caption{Evolution of the Questioner across training steps on the single-view relative direction task. Observations become increasingly detailed and spatially grounded, with generated questions more tightly coupled to visual evidence.}
\label{fig:solver_case}
\vspace{-1.2em}
\end{figure}

We present qualitative cases illustrating the evolution of the Questioner and Solver throughout training, using the single-view relative direction task as a representative example.

\paragraph{Questioner Evolution.}
As shown in Figure~\ref{fig:questioner_case}, the Questioner's output quality improves markedly across training steps. At Step 20 (Reward 0.1), the observation is brief and superficial, providing only coarse positional descriptions without meaningful spatial grounding; the generated question is thus weakly supported by the visual evidence. By Step 1116 (Reward 0.67), the observation becomes more structured, describing object sizes, foreground-background relationships, and relative positions with greater precision. At Step 1937 (Reward 1.0), the observation is detailed and spatially grounded, accurately localizing target objects within the scene and explicitly establishing the geometric relationship that motivates the question. This progression demonstrates that the Questioner progressively internalizes holistic scene perception and learns to generate questions that are tightly coupled to visual evidence.

\paragraph{Solver Evolution.}
As shown in Figure~\ref{fig:solver_case}, the Solver's reasoning process undergoes a parallel improvement. At Step 136 (Reward 0.1), the Solver produces a brief, unstructured description and arrives at an incorrect answer, failing to establish a coherent geometric reasoning chain. At Step 1116 (Reward 1.0), the reasoning becomes multi-step and logically structured, explicitly identifying object positions, inferring spatial relationships, and deriving the correct answer through sequential analysis. At Step 1860 (Reward 1.0), the Solver adopts a clearly enumerated reasoning format, decomposing the spatial inference problem into explicit sub-steps and arriving at a precise answer with consistent geometric justification. These cases collectively demonstrate that DGE-grounded self-evolution drives the internalization of structured spatial reasoning paradigms in both roles.

\end{document}